\documentclass[10pt,twocolumn,letterpaper]{article}

\usepackage{cvpr}              

\usepackage{graphicx}
\usepackage{amsmath}
\usepackage{amssymb}
\usepackage{booktabs}
\usepackage[normalem]{ulem}
\useunder{\uline}{\ul}{}
\usepackage{multirow}
\usepackage{makecell}
\usepackage{algorithm}
\usepackage{algpseudocode}
\usepackage[accsupp]{axessibility} 

\usepackage[pagebackref,breaklinks,colorlinks]{hyperref}

\usepackage[capitalize]{cleveref}
\crefname{section}{Sec.}{Secs.}
\Crefname{section}{Section}{Sections}
\Crefname{table}{Table}{Tables}
\crefname{table}{Tab.}{Tabs.}

\def\cvprPaperID{7269} 
\def\confName{CVPR}
\def\confYear{2023}

\begin{document}

\title{PosterLayout: A New Benchmark and Approach for Content-aware Visual-Textual Presentation Layout}

\author{HsiaoYuan Hsu\textsuperscript{1,2}, Xiangteng He\textsuperscript{1,2}, Yuxin Peng\textsuperscript{1,2}\thanks{Corresponding author.}, Hao Kong\textsuperscript{3} and Qing Zhang\textsuperscript{3}\\
\textsuperscript{1}Wangxuan Institute of Computer Technology, Peking University \textsuperscript{2}National Key Laboratory for \\Multimedia Information Processing, School of Computer Science, Peking University \textsuperscript{3}Meituan\\
{\tt\small kslh99@stu.pku.edu.cn, \{hexiangteng, pengyuxin, konghao\}@pku.edu.cn, zhangqing31@meituan.com}
}
\maketitle

\begin{abstract}
   Content-aware visual-textual presentation layout aims at arranging spatial space on the given canvas for pre-defined elements, including text, logo, and underlay, which is a key to automatic template-free creative graphic design.
   In practical applications, e.g., poster designs, the canvas is originally non-empty, and both inter-element relationships as well as inter-layer relationships should be concerned when generating a proper layout.
   A few recent works deal with them simultaneously, but they still suffer from poor graphic performance, such as a lack of layout variety or spatial non-alignment.
   Since content-aware visual-textual presentation layout is a novel task, we first construct a new dataset named \textbf{PKU PosterLayout}, which consists of 9,974 poster-layout pairs and 905 images, i.e., non-empty canvases. It is more challenging and useful for greater \textbf{layout variety}, \textbf{domain diversity}, and \textbf{content diversity}.
   Then, we propose design sequence formation (DSF) that reorganizes elements in layouts to imitate the design processes of human designers, and a novel CNN-LSTM-based conditional generative adversarial network (GAN) is presented to generate proper layouts.
   Specifically, the discriminator is design-sequence-aware and will supervise the "design" process of the generator.
   Experimental results verify the usefulness of the new benchmark and the effectiveness of the proposed approach, which achieves the best performance by generating suitable layouts for diverse canvases. 
   The dataset and the source code are available at \href{https://github.com/PKU-ICST-MIPL/PosterLayout-CVPR2023}{https://github.com/PKU-ICST-MIPL/PosterLayout-CVPR2023}.
\end{abstract}

\section{Introduction}
\label{sec:intro}

\begin{figure}[t]
  \vspace{0.2\baselineskip}  
  \centering
  \begin{tabular}{c@{\hspace{2pt}}c@{\hspace{2pt}}c}
    \multirow{2}{*}[16.2pt]{\includegraphics[width=0.14\textwidth] {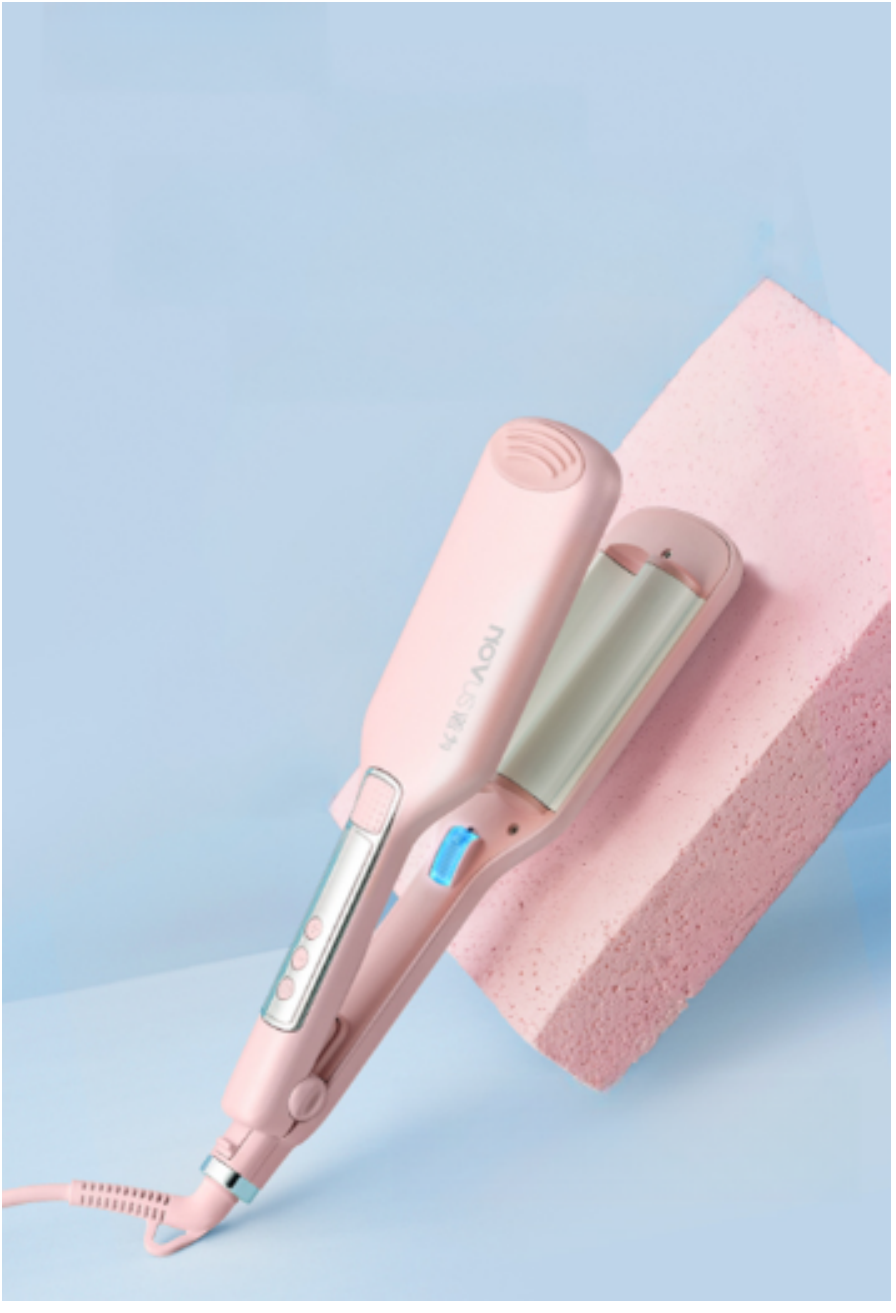}} &
    {\includegraphics[width=0.085\textwidth] {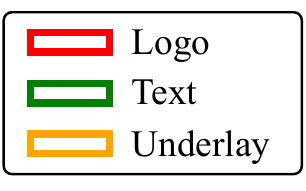}} &
    \multirow{2}{*}[16.2pt]{\includegraphics[width=0.14\textwidth] {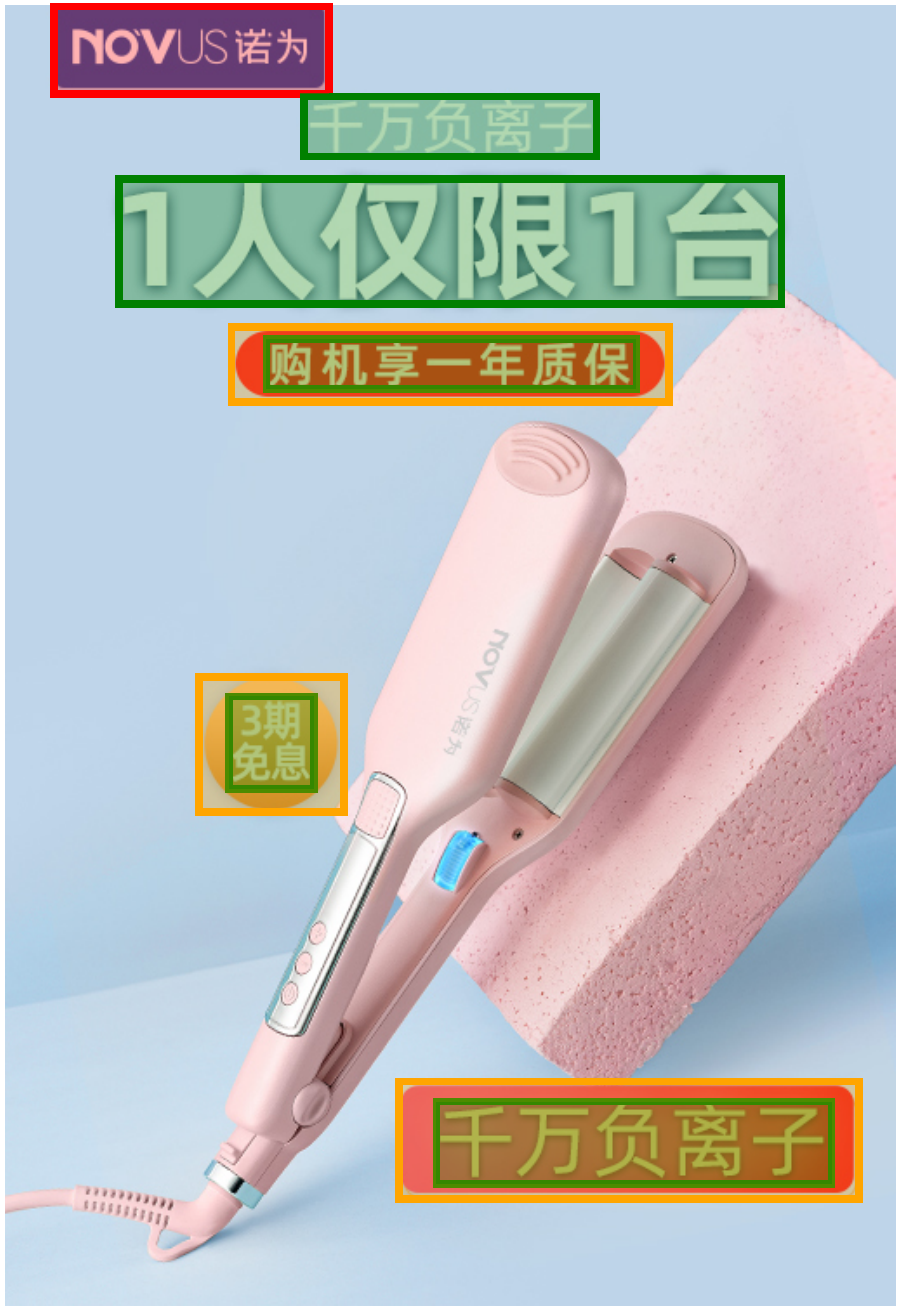}} \\
    & \includegraphics[width=0.1\textwidth] {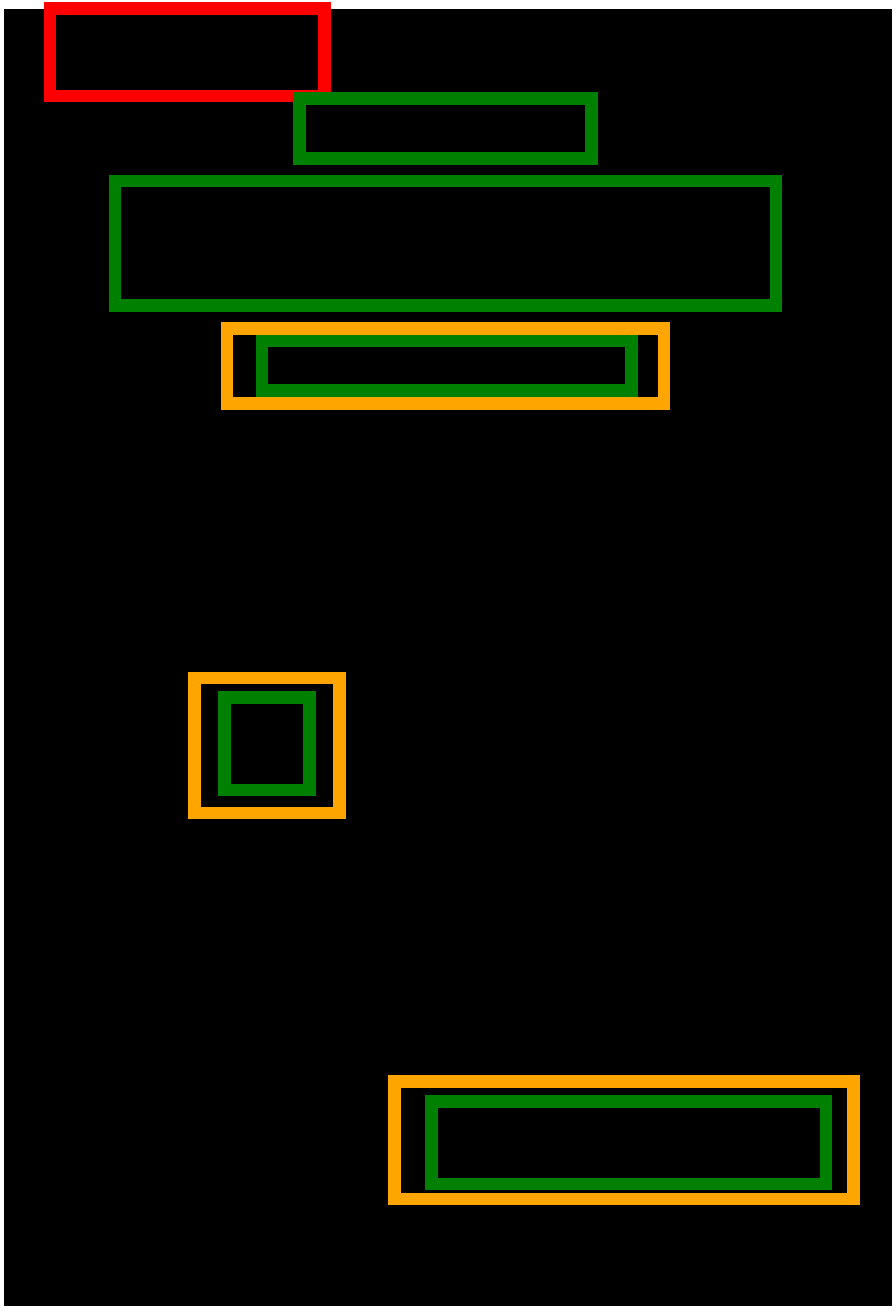} & \\
    (a) & (b) & (c)
  \end{tabular}
    \vspace{-0.4\baselineskip}
  \caption{Content-aware visual-textual presentation layout: (a) Non-empty canvas; (b) Content-aware layout; (c) An example of rendered presentation applying (b).}
  \label{fig:layout_example}
    \vspace{-1\baselineskip}  
\end{figure}

    Nowadays, visual-textual presentation rendering informative and decorative elements on an image, i.e., canvas, is widely used to convey information, such as advertisement posters \cite{Donovan-2015-CHI, Li-2020-TVCG, Guo-2021-CHI}, magazines \cite{Yang-2016-TOMM, Zheng-2019-TOG}, and so on \cite{Donovan-2014-TVCG, Lee-2020-ECCV, Guo-2021-SPML}. The basis of these creative works is the layout that indicates the spatial structure of the arranged elements, as shown in \cref{fig:layout_example}, which is also a key factor influencing their effectiveness and aesthetics.
    For their popularity and usefulness, not only experienced designers but also novice ones or "newbies" are commonly in need of creating them. People resort to pre-defined templates when they don't have enough prerequisites or need mass production.
    However, one can easily imagine that these templates harshly limit the flexibility and diversity of the presentations. These drawbacks of relying on templates hence highlight the importance and practicality of template-free creative graphic design, which can be preliminarily satisfied by automatically generating visual-textual presentation layouts.
    
    With the advance in deep learning and big data, more and more data-driven approaches for visual-textual presentation layout have emerged in this decade. 
    However, most of them have only been devoted to mining the relationship between elements and seldom concerned between layers, i.e., layout and canvas. Without proper constraints, elements are easily prone to cover the salient contents in the canvas, causing a severe occlusion problem. For example, in advertisement poster design, one of the most content-rich presentations, the product in the canvas shouldn't be over-occluded, which is no doubt.
    A few works \cite{Min-2022-IJCAI, Cao-2022-ACMMM} deal with inter-element and inter-layer relationships simultaneously, but they still suffer from poor graphic performance, such as a lack of layout variety or spatial non-alignment.
    To this end, we propose a CNN-LSTM-based generative adversarial network (GAN) conditioned by the input canvases to generate layouts, which has a balanced performance on both graphic and content-aware metrics.
    
    CNN-LSTM is proved effective in time series forecasting or behavior analysis tasks \cite{Mutegeki-2020-ICAIIC, Hsu-2021-ICEA}.
    To enable this time-sensitive model in layout generation, we propose design sequence formation (DSF) to generate design sequences that imitate the design processes of human designers.
    In particular, elements in layouts are reorganized to involve implicit temporal features, and less important ones can be discarded painlessly.
    It is in line with the logic of human-computer interaction logic \cite{Guo-2021-CHI} and has the potential to help train the LSTM model on a training set of size smaller than 20,000 \cite{Vinyals-2016-ICLR}.
    GAN is a generative model that contains a discriminator and a generator gaming against each other to learn the distribution of training data.
    In the proposed design sequence GAN (DS-GAN), the discriminator is design-sequence-aware and will supervise the "design"  process, i.e., generated layouts, of the generator under the constraints of the given canvas.
    As far as we know, this paper is the first adoption of CNN-LSTM in layout generation.
    
    Since content-aware visual-textual presentation layout remains a novel task, there is only one public dataset in the field, and it has insufficient variety.
    In this paper, we first construct and release a new dataset and benchmark named \textbf{PKU PosterLayout}, which consists of 9,974 poster-layout pairs and 905 images, i.e., non-empty canvases. Each layout is represented by a set of elements labeled with class and bounding box. We collect data from multiple sources to guarantee diversity and variety in content, domain, and layout, supporting it as a challenging benchmark expected to encourage further research.
    %
    %
    Besides the dataset, we propose and clearly define new metrics to accompany the old ones, a total of eight graphic and content-aware metrics. They evaluate the layouts in terms of utilization, non-occlusion, and aesthetics. 
    Both quantitative results and visualized results show that the proposed approach outperforms other approaches by generating proper layouts on diverse canvases.
    
    
    We summarize the contribution of this paper as follows:
    \begin{itemize}
        \item A new and more challenging dataset and benchmark for content-aware visual-textual presentation layout, \textbf{PKU PosterLayout}, consists of 9,974 poster-layout pairs and 905 images, with greater diversity and variety in content, domain, and layout.
        \item An algorithm for design sequence formation (\textbf{DSF}) converts plain layout data into design sequences involving temporal features by imitating the design process of human designers.
        \item A CNN-LSTM-based GAN, design sequence GAN (\textbf{DS-GAN}), is conditioned by images and learns the distribution of design sequences to generate content-aware visual-textual presentation layouts. It makes a good trade-off between graphic and content-aware metrics, which outperforms the other approaches.
    \end{itemize}

\section{Related Work}
\label{sec:relate_work}

Research on content-agnostic visual-textual presentation has developed for a relatively long time, assuming the given canvas is empty.
O’Donovan et al. \cite{Donovan-2014-TVCG} proposed an energy-based model that penalizes the part of layouts that violates pre-defined, complex design principles and thus could obtain a more desirable one after non-linear inverse optimization.
The authors further presented a system \cite{Donovan-2015-CHI} adopting this model with simpler principles, such as the size of elements and pair alignment, to alleviate time-consuming problem in heuristics.

Li et al. proposed LayoutGAN \cite{Li-2020-TPAMI}, taking a big step forward in data-driven approaches by introducing GANs in layout tasks.
It adopted a differentiable wireframe rendering layer flattening layouts and canvases into wireframe images, remaining the discrimination process an image classification problem.
In contrast, it differed from a conventional GAN in starting from a random initial layout that is primitively valid and modulating it into an eligible one instead of synthesizing layouts from fully random noise.
The authors further presented an attribute-conditioned LayoutGAN \cite{Li-2020-TVCG} that guides the layout with the given element attributes, such as minimum size, fixed aspect ratio, and reading order of elements.
Moreover, it accompanied elements dropout in the discrimination process, forcing the discriminator to be aware of the local pattern of layouts, which is helpful in visual-textual presentation layout.
Besides the element attributes, Zheng et al. \cite{Zheng-2019-TOG} demonstrated the efficiency of concerning the visual and textual semantics of the elements and presentation topics. They proposed an embedding network fusing cross-modal features to condition the GAN.

\begin{table*}[t]
\centering
\resizebox{\linewidth}{!}{
\begin{tabular}{|l|l|ll|l|l|l|l|}
\hline
 &
  \multicolumn{1}{c|}{\multirow{2}{*}{Status}} &
  \multicolumn{2}{c|}{\# data} &
  \multicolumn{2}{c|}{Layout} &
  \multicolumn{1}{c|}{\multirow{2}{*}{Canvas}} &
  \multicolumn{1}{c|}{\multirow{2}{*}{Content category}}
  \\ \cline{3-6}
   &         & \multicolumn{1}{c|}{Layout}  & \multicolumn{1}{c|}{Canvas} & \multicolumn{1}{c|}{Element types} & \multicolumn{1}{c|}{Complex?} & \multicolumn{1}{c|}{} & \multicolumn{1}{c|}{}
   \\ \hline
\multirow{2}{*}{NDN \cite{Lee-2020-ECCV}} &
  \multirow{2}{*}{Private} &
  \multicolumn{1}{l|}{\multirow{2}{*}{500}} &
  \multirow{2}{*}{NaN} &
  Text, logo, RoI, image, &
  \multirow{2}{*}{No} &
  \multirow{2}{*}{Empty} &
  \multirow{2}{*}{Car}
  \\
   &         & \multicolumn{1}{l|}{}        &        & brand name, button    &                       &             &                         
   \\ \hline
    ICVT \cite{Cao-2022-ACMMM} & 
    Private & 
    \multicolumn{1}{l|}{117,624} & 
    166    & 
    Text, logo, underlay  & 
    Yes &
    Non-empty             &
    Not given              
\\ \hline
\multirow{2}{*}{CGL-GAN \cite{Min-2022-IJCAI}} &
  \multirow{2}{*}{Public} &
  \multicolumn{1}{l|}{\multirow{2}{*}{60,548}} &
  \multirow{2}{*}{1,000} &
  Text, logo, underlay, &
  \multirow{2}{*}{No} &
  \multirow{2}{*}{Non-empty} &
  Cosmetics, electronics,
  \\
                           &         & \multicolumn{1}{l|}{}        &        & embellishment         &              &         & clothing, etc.                     
\\ \hline
\multirow{3}{*}{
    \begin{tabular}{@{\hspace{-2pt}}l}
         PKU PosterLayout\\
         (Ours)
    \end{tabular}
} &
  \multirow{3}{*}{Public} &
  \multicolumn{1}{l|}{\multirow{3}{*}{9,974}} &
  \multirow{3}{*}{905} &
  \multirow{3}{*}{Text, logo, underlay} &
  \multirow{3}{*}{Yes} &
  \multirow{3}{*}{Non-empty} &
  Cosmetics, electronics, \\
   &         & \multicolumn{1}{l|}{}        &        &        &               &                       & clothing,  delicatessen, \\ 
                              &         & \multicolumn{1}{l|}{}        &        &                       &           &            & toys/instruments, etc.   
    \\ \hline
\end{tabular}
}
\caption{Comparison of properties of benchmark for visual-textual presentation layout.}
\label{tab:com_benchmark}
\end{table*}

\begin{figure*}[h]
  \centering
    \begin{tabular}{c@{\hspace{12pt}}c@{\hspace{12pt}}c@{\hspace{12pt}}c}
    \multirow{3}{*}[76.65pt]{\includegraphics[width=0.41\textwidth] {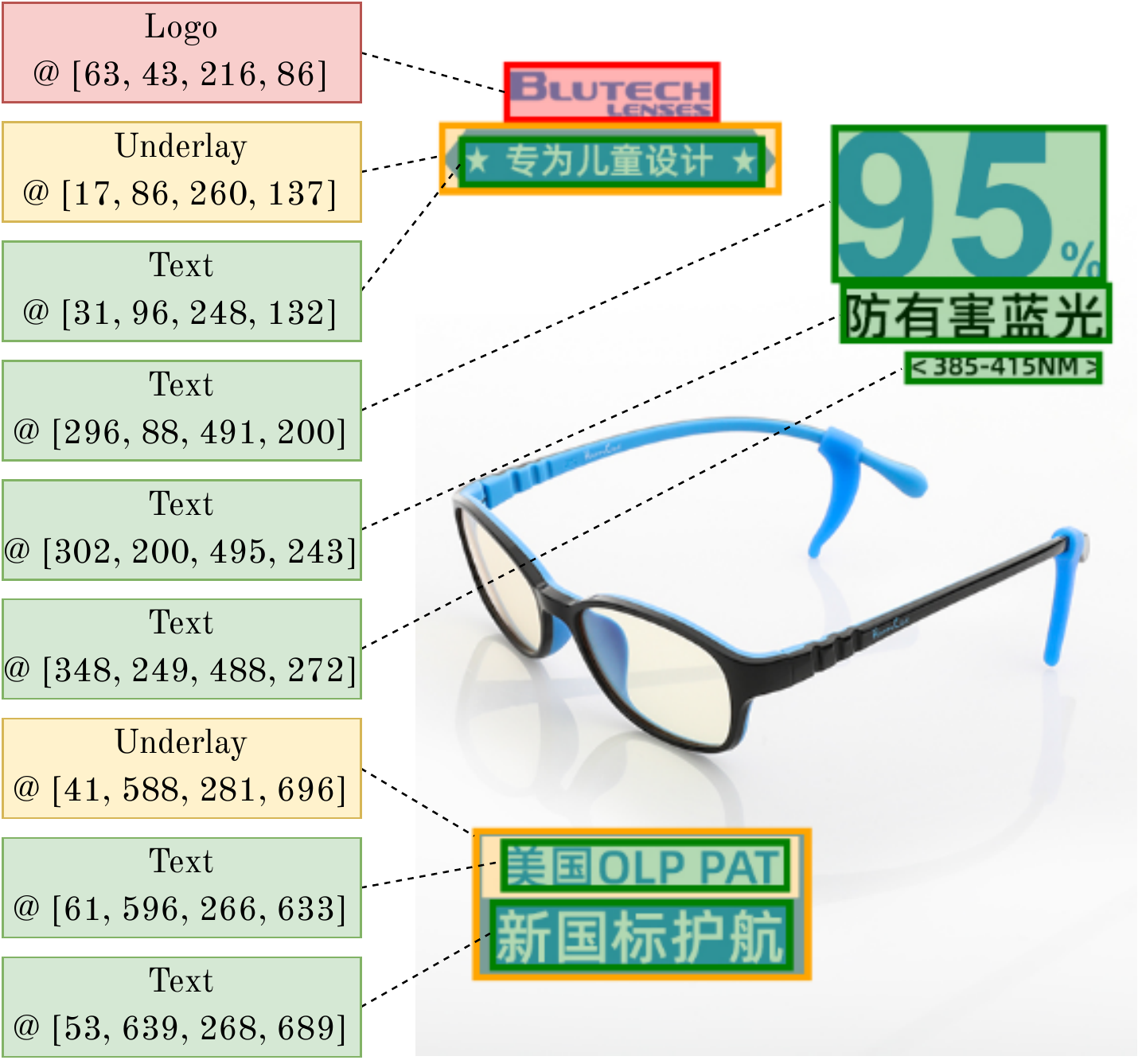}} &
    \includegraphics[width=0.12\textwidth] {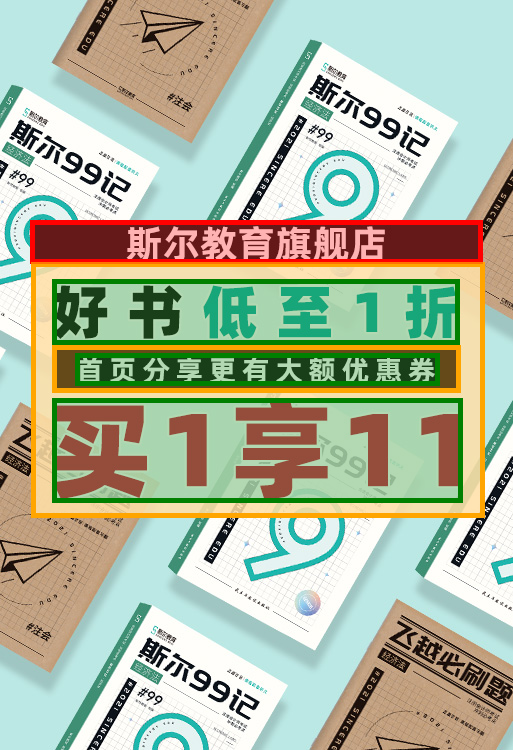} &
    \includegraphics[width=0.12\textwidth] {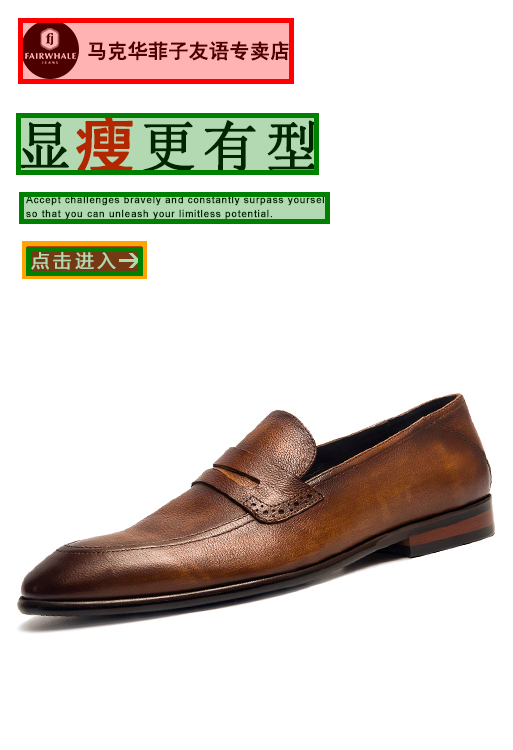} &
    \includegraphics[width=0.12\textwidth] {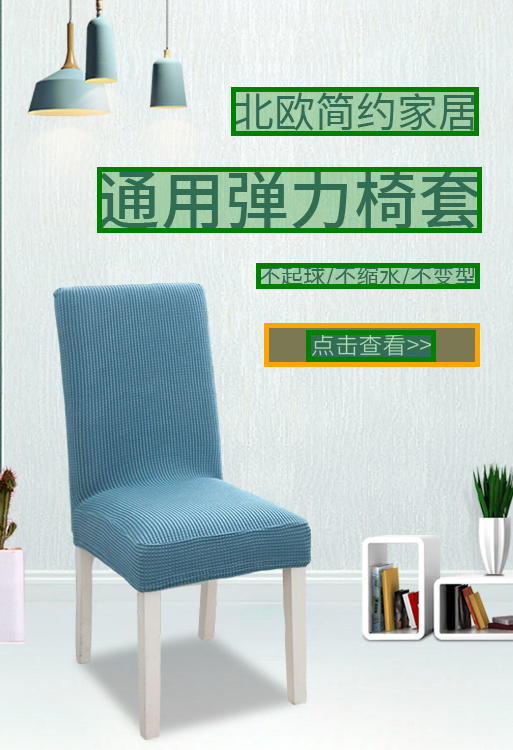}
    \\
    & (b) Middle & (c) Top-left & (d) Top-right \\
     &
    \includegraphics[width=0.12\textwidth] {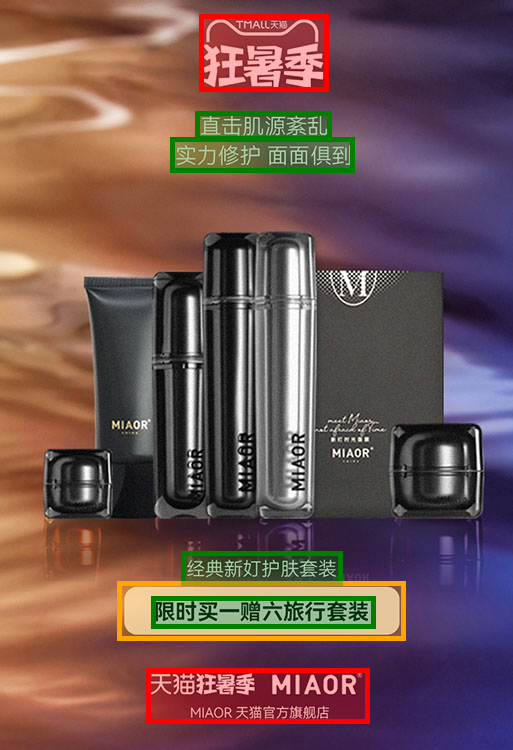} &
    \includegraphics[width=0.12\textwidth] {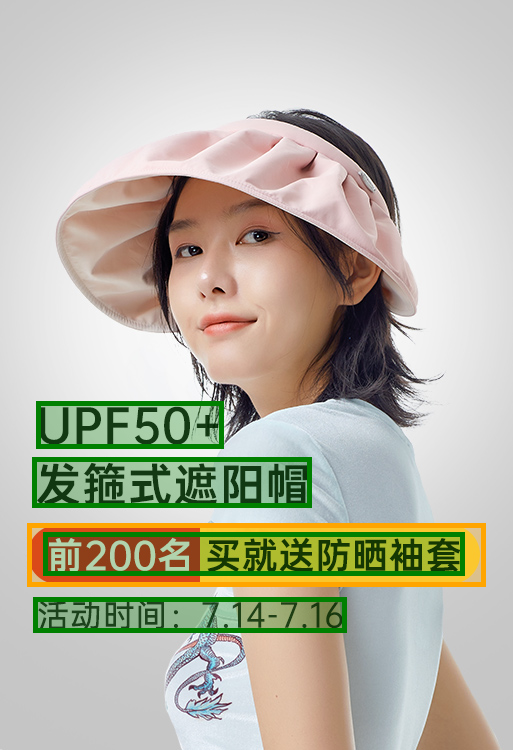} & 
    \includegraphics[width=0.12\textwidth] {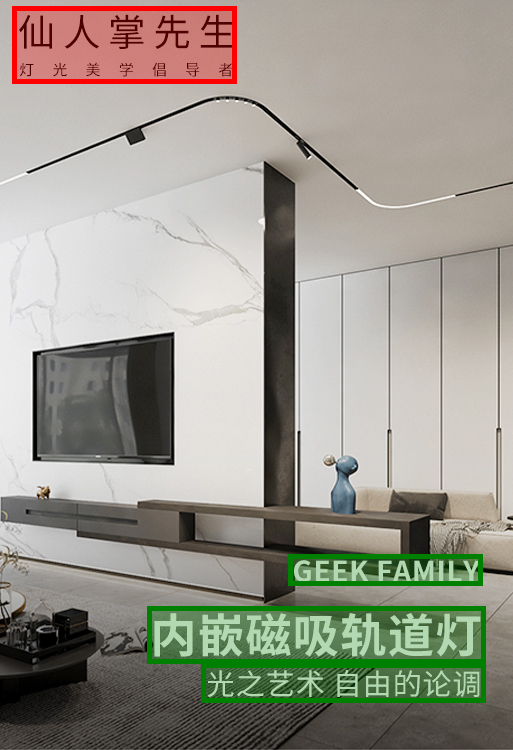} \\
    (a) & (e) Top-bottom & (f) Bottom-left & (g) Bottom-right
    \end{tabular}
  \caption{Examples of poster-layout pairs in PKU PosterLayout.}
  \label{fig:dataset_example}
    \vspace{-1\baselineskip}  
\end{figure*}

Kikuchi et al. proposed LayoutGAN++ \cite{Kikuchi-2021-ACMMM} demonstrating an improvement in handling user-specific constraints by optimizing layout in latent space.
It got rid of using wireframe images with respect to the findings that the rendering layer is unstable with a dataset of a limited size.
Similarly, Lee et al. \cite{Lee-2020-ECCV} were concerned with user-specific constraints and dealt with them using a graph neural network modeling elements as nodes and their relationships as edges.
Clarification is needed that these user-specific constraints are merely inter-layout and insufficient for the task interested in this paper.
Specifically, content-aware visual-textual presentation layout concerns both inter-layout and inter-layer relationships, i.e., layout and canvas, which is driven by canvas with no mandatory constraints attached.
However, the ideas behind these content-agnostic approaches are still worthy of exploring to enhance the performance of content-aware ones.

For content-aware visual-textual presentation layout, Zhuo et al. proposed CGL-GAN \cite{Min-2022-IJCAI} utilizing the standard transformer block, which is the first and most relevant work in the disciplines.
The encoder and decoder had the visual features of canvas and the embedding of layout as input, respectively. Therefore, the self-attention and cross-attention in the decoder can simultaneously model the inter-layout and inter-layer relationships. 
Although experimental results demonstrated its usefulness in improving content-aware metrics, it had a relatively poor performance in graphic metrics, especially the spatial non-alignment.
Following almost the same ideas as CGL-GAN, most recently, Cao et al. \cite{Cao-2022-ACMMM} proposed an image-conditioned variational transformer with the proposed geometry-aligned fusion formula applied in the cross-attention layer.
Not surprisingly, it suffered from a similar problem, especially the undesired overlap between elements.
What these approaches encounter encourages us to research and propose a novel approach making a trade-off between two types of metrics to generate the most proper layout.


\section{A New Benchmark: PKU PosterLayout}
\label{sec:posterlayout}

A few datasets related to visual-textual presentation layouts for posters have been presented in previous works.
\cref{tab:com_benchmark} shows the properties of these datasets.
In detail, the pre-defined element types, target canvas, scale, i.e., amounts of layout and canvas data, and diversity, i.e., categories of posters, are compared.
Since some datasets are still private to this day, statistics are from their source papers.
NDN \cite{Lee-2020-ECCV} presented a banner-layout dataset composed of 500 car advertisements for validating its content-agnostic approach, which doesn't completely align with the target task and suffers from scale problems and a lack of diversity.
ICVT \cite{Cao-2022-ACMMM} presented a large-scale 117k poster-layout dataset, strongly supporting training content-aware approaches. However, its usefulness for validation is doubted since the excessively small testing set, i.e., 166 canvases, and the unknown diversity.
CGL-GAN \cite{Min-2022-IJCAI} presented a dataset with 60k poster-layout pairs and 1k canvases. It seems to be overall satisfactory; however, it collects all data from a single source, covers only a few categories with a disproportionate proportion, and does not involve complex layouts containing more than 10 elements. These drawbacks limit its generality.

\begin{figure}[t]
  \vspace{-0.4\baselineskip}  
  \centering
  \includegraphics[width=0.45\textwidth] {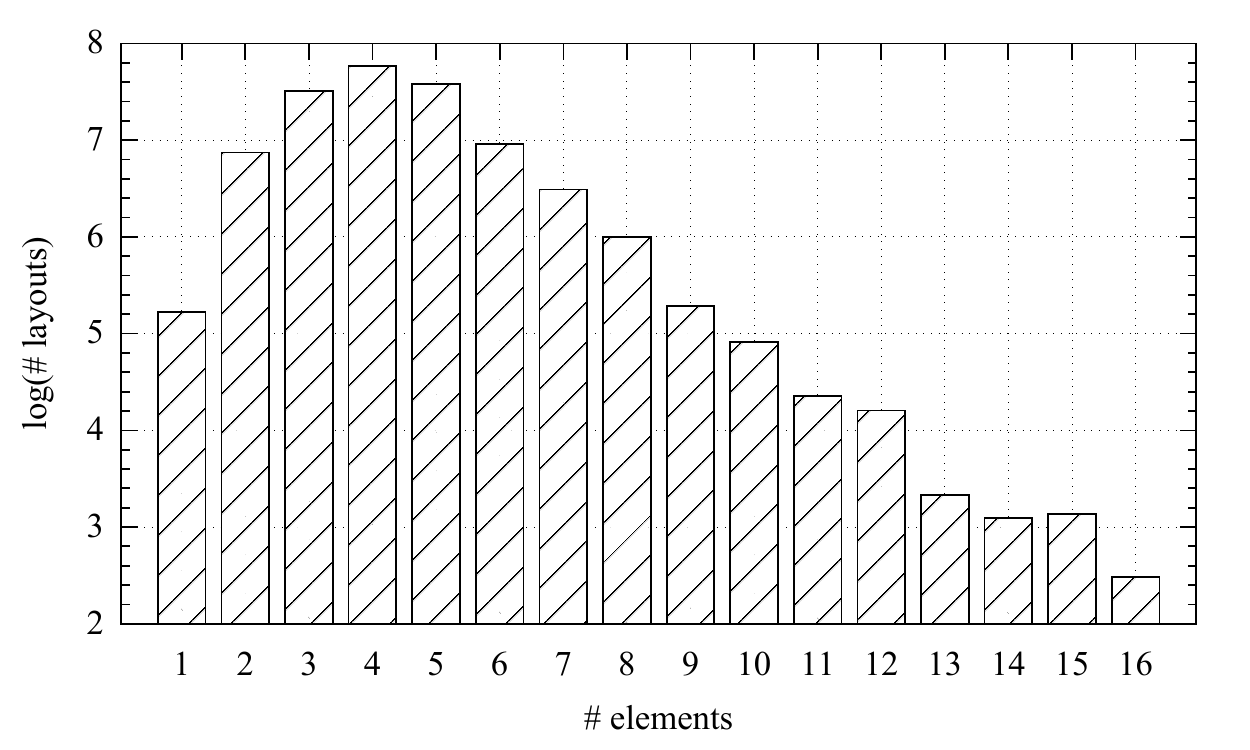}
  \caption{Statistics on layout variety in PKU PosterLayout.}
  \label{fig:htg_ds}
    \vspace{-1\baselineskip}  
\end{figure}

\begin{figure*}[t]
  \centering
  \includegraphics[width=0.952\textwidth] {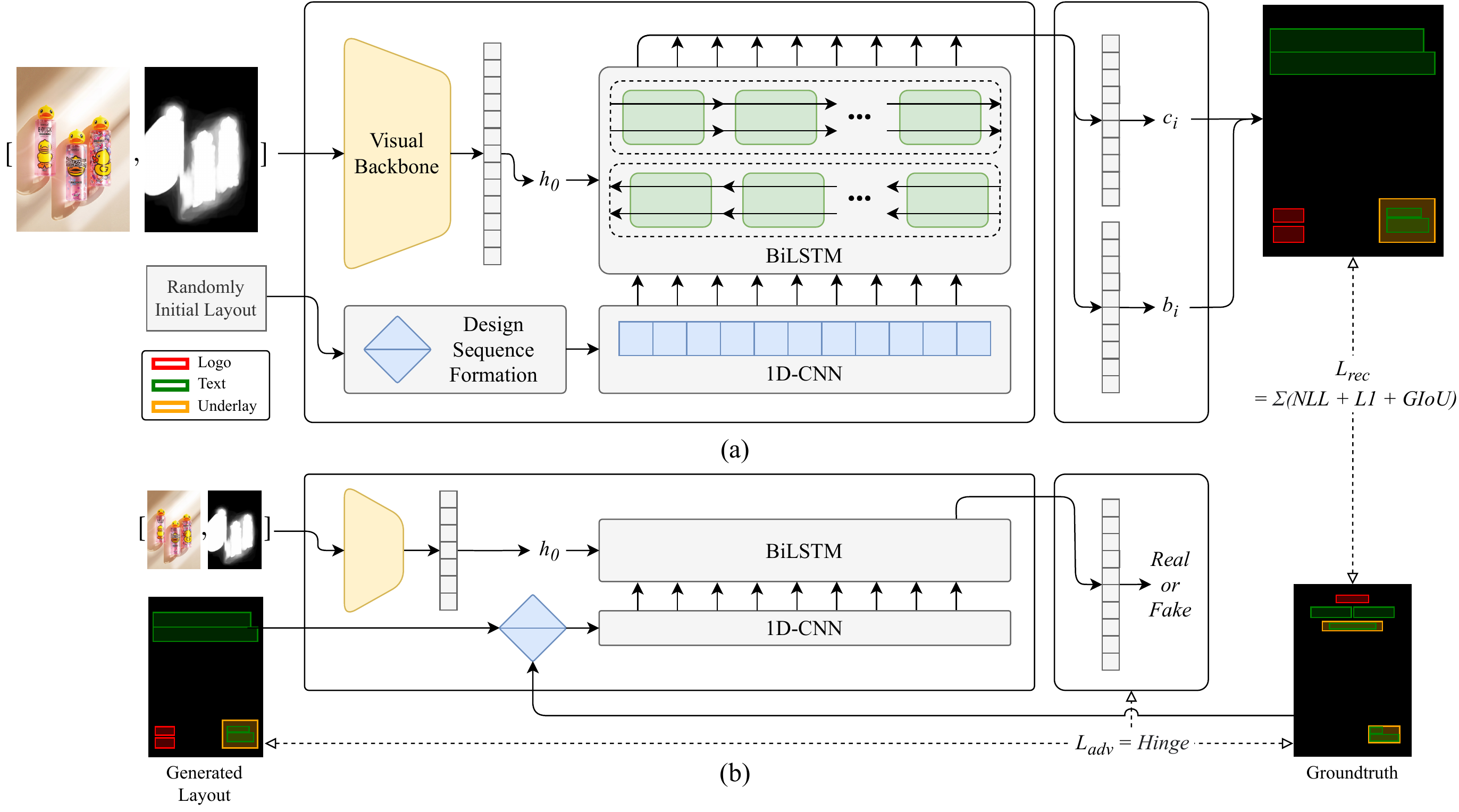}
  \caption{An overview of the proposed approach: (a) generator of DS-GAN; (b) discriminator of DS-GAN.}
  \label{fig:overview}
    \vspace{-1\baselineskip}  
\end{figure*}

To this end, a new dataset and benchmark, PKU PosterLayout, is constructed in this paper. It shows advantages in domain diversity, content diversity, and layout variety.
We first adopt posters from a subset of an e-commerce posters dataset \cite{Jiang-2022-ACMMM-Erase} and define element types.
Each poster is paired with a layout composed of a variable-length set \(n\) of elements, \(L=\{e_{i}\ |\ i=0, 1, ..., n-1\}\). Each element \(e\) is represented with its type \(c\) and bounding box \(b=[x_1, y_1, x_2, y_2]\), standing for the top-left and bottom-down coordinates. 
Without loss of generality, three element types are defined, including text, logo, and underlay, as shown in \cref{fig:dataset_example}.
Logos are graphic elements that indicate brand names or promotion activities, and underlays are decorations below or around any elements.
While underlay-below-underlay is allowed, a strict rule to follow when labeling is that every underlay must decorate at least one text or logo independently, as shown in \cref{fig:dataset_example}(b).
Finally, texts are all the other informative elements that are not logos.
Manually labeling from scratch is inefficient and infeasible, and thus an iterative scheme is applied with the help of an object detection model \cite{Girshick-2015-ICCV-FRCNN}.
All poster-layout pairs are refined and examined to be correct by human annotators, and then the rendered elements on posters are erased using a Fourier-convolution-based inpainting method \cite{Suvorov-2022-WCAC-LaMa}.
\cref{fig:htg_ds} shows statistics on layout variety observed briefly from the number of elements in layouts, indicating a broad distribution. It is emphasized that there are several complex layouts with more than 10 elements in PKU PosterLayout. Therefore, it can be sufficient for layout tasks under complicated settings, such as \cite{Jiang-2022-AAAI-CTF}, while \cite{Min-2022-IJCAI} cannot.

Afterward, we collect background and product images via searches to create canvases of various qualities while carefully keeping the numbers of ones even in each category. There are totally nine categories, including food/drinks, cosmetics/accessories, electronics/office supplies, toys/instruments, clothing, sports/transportation, groceries, appliances/decor, and fresh produce.
Eventually, 9,974 poster-layout pairs and 905 canvases constitute PKU PosterLayout, a median-scale dataset and benchmark with guaranteed layout variety, domain diversity, and content diversity.

\section{Proposed Approach}
\label{sec:approach}
After analyzing the weaknesses of previous works, a novel generative model that makes a good trade-off between graphic and content-aware performance is proposed in this paper.
This section will start with the basic ideas of design sequences and the algorithm used to form them automatically.
Afterward, the proposed CNN-LSTM-based GAN empowered by these sequences will be expounded.
An overview of the proposed approach is shown in \cref{fig:overview}.

\begin{figure}[t]
  \centering
  \includegraphics[width=0.4\textwidth] {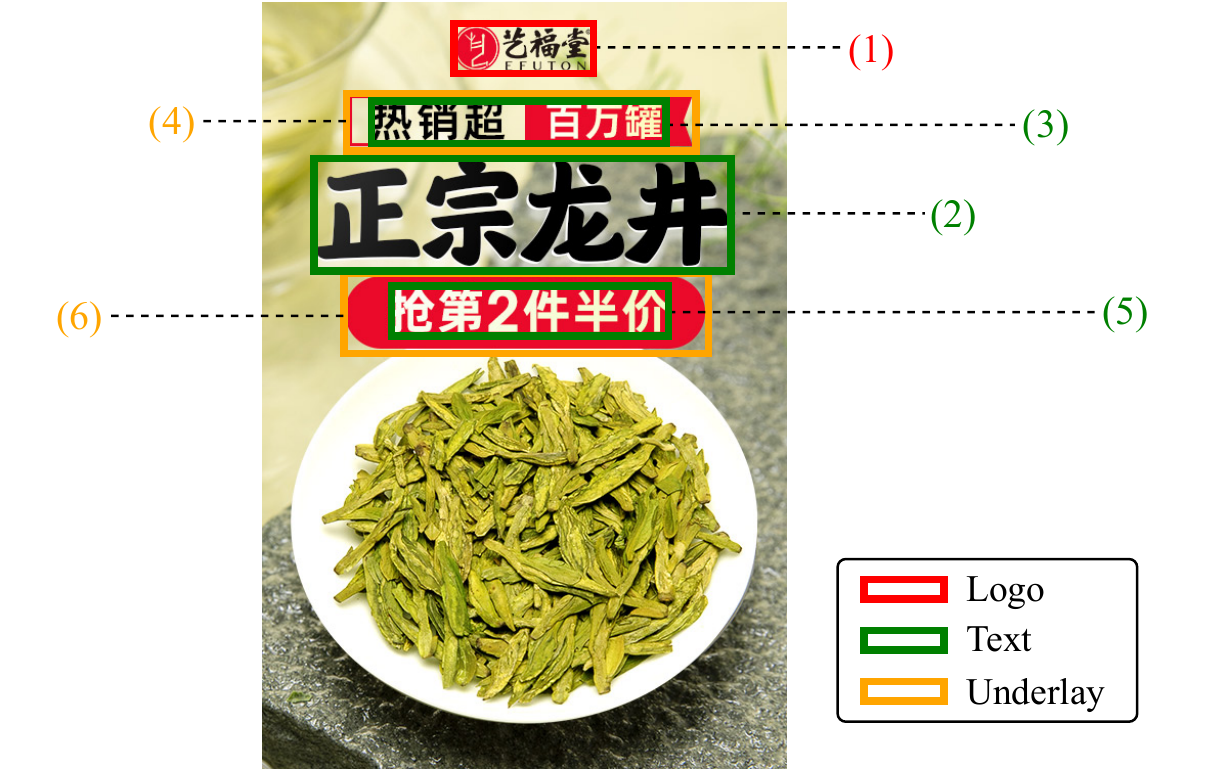}
  \caption{A design sequence formed by the proposed DSF.}
  \label{fig:dsf}
    \vspace{-1\baselineskip}  
\end{figure}

\begin{algorithm}[b]
\caption{Algorithm for design sequence formation}
\label{alg:dsf}
\begin{algorithmic}[1]
\Require Layout $ L=\{e_{i}\ |\ i=0, 1, ..., n-1\} $
\Ensure $R$ = design sequence formed from $L$
\State $R \gets \{\}$
\State $l \gets \{e_{i} \ |\ c_{i}$ is \emph{logo}$\}$
\State $t \gets \{e_{i} \ |\ c_{i}$ is \emph{text}$\}$
\State $u \gets \{e_{i} \ |\ c_{i}$ is \emph{underlay}$\}$
\State Sort $(l, (y_{top}, x_{left}))$ in ascending order
\State Sort $(t, area)$ in descending order
\State $G \gets$ groups of $l$ and $t$ with commonly overlaid $u$
\State \Comment{Stable merging}
\State Queue $Inst \gets Concat(l, t)$
\While{$Inst \neq \emptyset$}
\State $Inst' \gets pop(Inst)$
\If{$Inst' \notin R$}
\State $G' \gets G \supset Inst'$
\State $u' \gets u$ overlaid $G'$
\State $push(R, G')$
\State $push(R, u')$
\EndIf
\EndWhile
\end{algorithmic}
\end{algorithm}

\subsection{Design Sequence Formation}

Referring to Guo et al.'s work studying artistic creation in human-computer interaction \cite{Guo-2021-CHI}, modeling human designers' behaviors can be a promising way toward content-aware visual-textual presentation layout.
Specifically, behaviors are represented by design sequences, which indicate the order human designers place elements on canvases.
Since the layout data is plain and short of this information, an algorithm for DSF is presented, shown in \cref{alg:dsf}.
The main principle of DSF is putting the most informative, significant elements at the front and vice versa, which is a basic understanding of the design process.
Following a rule of thumb, the conspicuousness of logos is influenced by reading order, e.g., from top-left to bottom-right \cite{Li-2020-TVCG}, and thus their top-left coordinates are chosen as a criterion.
For texts, their areas are chosen, which is obvious.
Underlays, as the name implies, must be below others and thus won't be put into the sequence until all elements overlaying it get arranged.
\cref{fig:dsf} gives an example of a design sequence formed by DSF.
By maintaining the descending order of element importance, design sequences will not be severely disrupted even if the last few elements are discarded.
This additional benefit enhances the CNN-LSTM model adopted later, for it takes a fixed-length input.

\subsection{Design Sequence GAN}
\label{sec:dsgan}
GAN is a generative model that contains two sub-models, a discriminator and a generator.
The former is devoted to creating fake samples that sufficiently cheat the latter, and the latter is devoted to making a conscious distinction between the real and the fake samples.
As time goes by, they keep gaming against each other and can learn the distribution of training data.
In this paper, a CNN-LSTM-based GAN is proposed.
CNN-LSTM is a hybrid model well-known for its potential to handle time-series forecasting and behavior analysis \cite{Mutegeki-2020-ICAIIC, Hsu-2021-ICEA}.
With the help of the DSF presented above, layout generation is transferred to be a novel behavior modeling problem and thus becomes solvable for CNN-LSTM.

\begin{table*}[t]
\centering
\begin{tabular}{|l|c|c|c|c|c|c|c|c|c|}
\hline
 &
  Target &
  $Val \uparrow$ &
  $Ove \downarrow$ &
  $Ali \downarrow$ &
  $Und_{l} \uparrow$ &
  $Und_{s} \uparrow$ &
  $Uti \uparrow$ &
  $Occ \downarrow$ &
  $Rea \downarrow$ \\ \hline
SmartText \cite{Li-2021-TMM} & $T$  & - & -      & -      & -      & -      & 0.0849 & \textbf{0.0912} & \textbf{0.1528} \\ \hline
CGL-GAN \cite{Min-2022-IJCAI}       & $V$-$T$ & 0.7066 & 0.0605 & 0.0062 & \textbf{0.8624} & 0.4043 & 0.2257 & 0.1546          & 0.1715          \\ \hline
DS-GAN (Ours) &
  $V$-$T$ &
  \textbf{0.8788} &
  \textbf{0.0220} &
  \textbf{0.0046} &
  0.8315 &
  \textbf{0.4320} &
  \textbf{0.2541} &
  0.2088 & 
  0.1874 \\ \hline
\end{tabular}
\caption{Comparison of quantitative results.}
\label{tab:results}
\vspace{-1\baselineskip} 
\end{table*}

Even more sensible, another reason to use LSTM is that it is triggered by an initial hidden state, exactly where the conditions can be attached.
Specifically, instead of roughly concatenating visual and layout embedding, like what conditional GANs do to make models image-conditioned, DS-GAN perceives visual content by explicitly initializing its hidden state $h_0$ with visual features $F$, as
\begin{equation}
 \label{eq:visual_f}
 \begin{gathered}
F = \text{ResNet-FPN}([I, max(S_{\text{PFPN}}, S_{\text{BASNet'}})]) \\
h_0 = Linear(F),
 \end{gathered}
\end{equation}
where $I$ represents an input image, i.e., repaired poster or canvas,  $S_{i}$ represents a saliency map constructed by different domain salient detection methods \cite{Wang-2020-AAAI-PFPN, Zhang-2020-ICME}, $[f_1, f_2]$ represents feature concatenation along channel-axis, operator ResNet-FPN refers to that presented in \cite{Min-2022-IJCAI}, operator $max$ represents a pixel-wise maximum operation, and operator $Linear$ represents a fully connected layer.
Then, design sequences formed from randomly initialized layouts \cite{Li-2020-TPAMI} (in the generator) or real/fake samples (in the discriminator) are directly input to the CNN-LSTM model, with no extra embedding layers needed.

Both generator and discriminator are built with the structure expounded above, i.e., a CNN-LSTM model that takes visual features as the initial hidden state and design sequences as input.
Though, of course, they have different ending parts for their respective target.
In the generator, two fully connected layers are further cascaded to decode the output of the CNN-LSTM model into the type and bounding box of each element in generated design sequence, turning valid elements into a layout.
Besides the adversarial loss $L_{adv}$ discussed later, a reconstruction loss $L_{rec}$ \cite{Carion-2020-ECCV-DETR} aggregating negative log-likelihood (NLL) loss, L1 loss, and generalized intersection of union (IoU) loss helps to train the generator.
By contrast, the discriminator ends up with only one fully connected layer that classifies whether input design sequences are real samples.
The adversarial loss that guides the discriminator and, thus, the generator is the hinge loss.


\section{Experiment}
\label{sec:experiment}

For validating the proposed approach, quantitative indicators of content-aware visual-textual presentation layout are elaborated.
Then, several experiments based on the proposed benchmark, PKU PosterLayout, are conducted, enabling the comparison between existing approaches and the proposed one.
Results show that DS-GAN with DSF achieves the best performance by generating the most proper layouts on diverse canvases while making a good trade-off between the two aspects, i.e., graphic and content-aware metrics.

\subsection{Evaluation Metrics}
Eight metrics in two aspects are defined as follows. Some of them are newly proposed or clearly defined for the first time, including $Val \uparrow$, $Und \uparrow$, and $Occ \downarrow$. The up arrow indicates a higher value is better, and vice versa.

\textbf{Graphic metrics}
Validity, annotated as $Val \uparrow$, is the ratio of valid elements to all elements in the layout, where the area within the canvas of a valid element must be greater than 0.1\% of the canvas.
Note that all remaining metrics consider only valid elements.
Overlay, annotated as $Ove \downarrow$, is the average IoU of all pairs of elements except for underlay.
\emph{Non}-alignment, annotated as $Ali \downarrow$, is the extent of spatial non-alignment between elements, referring to \cite{Li-2020-TVCG}.
Underlay effectiveness, annotated as $Und \uparrow$, is the ratio of valid underlay elements to total underlay elements, where a valid one $u$ must truly decorate at least one non-underlay element $Inst$.
Subscript $l$ means loose, calculating
\(\frac{area(u \cap Inst)}{area(Inst)}\)
for each pair, and $Und_l$ takes the maximum value.
Subscript s means strict, scoring the underlay as 1 if there is a non-underlay element completely inside, otherwise, 0, and $Und_s$ takes the average score.

\textbf{Content-aware metrics}
Utility, annotated as $Uti \uparrow$, is the utilization rate of space suitable for arranging elements, implemented by the negative image $S'$ of the compounded saliency map $S$, as mentioned in \cref{sec:dsgan}.
In particular, the denominator is the total pixel values of $S'$, and the numerator is that of $S'$ masked all areas without elements.
In opposite, occlusion, annotated as $Occ \downarrow$, is the average pixel value of areas covered by elements in $S$.
\emph{Un}readability, annotated as $Rea \downarrow$, is the non-flatness of regions that text elements are solely put on, referring to CGL-GAN.

\subsection{Implementation Details}
Considering the complexity of respective tasks, the generator of DS-GAN is with ResNet50 backbone and 4-layer CNN-BiLSTM, while the discriminator is with ResNet18 backbone and 2-layer CNN-BiLSTM.
When training the network, layout data $c$ is in one-hot vector form, $b$ is in \([x_c, y_c, w, h]\) form, standing for center coordinates and the width, height of the bounding box, and the batch size is 128.
The weights in reconstruction loss are 2, 5, and 2 for NLL, L1, and generalized IoU loss, respectively.
The weight of reconstruction is constantly 1, and that of adversarial loss increases linearly from 0 to 1 in a warm-up of 100 epochs.
The entire network is trained for 300 epochs.
Adam optimizers are used, of which learning rates are initialized as: $10^{-4}$, $10^{-5}$ for the generator and its visual backbone, and $10^{-3}$, $10^{-4}$ for discriminator and its backbone. All experiments are carried out with Pytorch framework and four NVIDIA A100-SXM4-80GB GPUs.

\begin{table*}[h]
\centering
\begin{tabular}{|l|c|c|c|c|c|c|c|c|}
\hline
  &
  $Val \uparrow$ &
  $Ove   \downarrow$ &
  $Ali   \downarrow$ &
  $Und_{l} \uparrow$ &
  $Und_{s} \uparrow$ &
  $Uti \uparrow$ &
  $Occ \downarrow$ &
  $Rea \downarrow$
   \\ \hline
    without CNN-LSTM   & 0.6765 & 0.0888 & 0.0112 & 0.0106 & 0.0000 & 0.2155 & 0.2804 & 0.2015 \\ \hline
    with CNN-LSTM (DS-GAN) &
  \textbf{0.8788} &
  \textbf{0.0220} &
  \textbf{0.0046} &
  \textbf{0.8315} &
  \textbf{0.4320} &
  \textbf{0.2541} &
  \textbf{0.2088} & 
  \textbf{0.1874} \\ \hline
\end{tabular}
\caption{Ablation study on CNN-LSTM model.}
\label{tab:ablation_body}
\end{table*}

\begin{table*}[t]
\centering
\begin{tabular}{|l|c|c|c|c|c|c|c|c|c|}
\hline
 &
  $Val \uparrow$ &
  $Ove   \downarrow$ &
  $Ali   \downarrow$ &
  $Und_{l} \uparrow$ &
  $Und_{s} \uparrow$ &
  $Uti \uparrow$ &
  $Occ \downarrow$ &
  $Rea \downarrow$ &
  $AE \downarrow$
   \\ \hline
\multirow{2}{*}{Random} & \textbf{1.0000} & 0.0881    & 0.0062    & 0.7417    & 0.3243    & 0.2240    & {\ul 0.2475} & \textbf{0.1909} & 
\multirow{2}{*}{0.5730} \\
                        & \small{(+0.1454)}       & \small{(+0.0666)} & \small{(+0.0007)} & \small{(-0.1380)} & \small{(-0.1499)} & \small{(-0.0328)} & \small{(+0.0361)}    & \small{(+0.0035)}       &           \\ \hline
\multirow{2}{*}{Geometric} &
  {\ul 0.9667} &
  \textbf{0.0261} &
  {\ul 0.0050} &
  {\ul 0.7849} &
  {\ul 0.4433} &
  {\ul 0.2439} &
  0.2482 &
  0.1937 &
  \multirow{2}{*}{\ul 0.3486}
   \\
                        & \small{(+0.1215)}       & \small{(+0.0026)} & \small{(+0.0004)} & \small{(-0.0824)} & \small{(-0.0757)} & \small{(-0.0170)} & \small{(+0.0438)}    & \small{(+0.0052)}       &     \\ \hline
DSF-based &
  0.9572 &
  {\ul 0.0362} &
  \textbf{0.0043} &
  \textbf{0.8850} &
  \textbf{0.5824} &
  \textbf{0.2526} &
  \textbf{0.2341} &
  {\ul 0.1910} & \multirow{2}{*}{\textbf{0.3272}}
   \\
   (DS-GAN-8)
                        & \small{(+0.0784)}       & \small{(+0.0142)} & \small{(-0.0003)} & \small{(+0.0535)} & \small{(+0.1504)} & \small{(-0.0015)} & \small{(+0.0253)}    & \small{(+0.0036)}       &  \\ \hline
\end{tabular}
\caption{Ablation study on design sequence formation.}
\vspace{-1\baselineskip}
\label{tab:ablabtion}
\end{table*}

\subsection{Comparison with State-of-the-arts}
The proposed approach is compared with SmartText \cite{Li-2021-TMM} and CGL-GAN \cite{Min-2022-IJCAI}.
While the latter was mentioned in \cref{sec:relate_work}, the former was not, for it was for content-aware textual presentation layout.
Besides the lack of comparable existing approaches, we choose SmartText for two reasons.
First, based on a saliency-aware region proposal, it intrinsically experts in avoiding unreadability and occlusion, so a comparison is worthy to see how long the target task still has to go.
Second, however, SmartText suffers from a fatal problem in that it puts all elements into the selected anchor box in a rigid way, making graphic metrics purposeless.
It may be sufficient for textual presentation but not for visual-textual one, showing the difference between them and the importance of the novel task discussed in this paper.
%

In the experiments, 905 canvases from nine categories in PKU PosterLayout are used for validation.
For fairness, in all approaches, the length of each design sequence (or just plain layout) is the maximum number of elements in all layout data, which means non-objects are padded if necessary.
We leave the effects of discarding less significant elements in the ablation study.
\cref{tab:results} shows the quantitative results of the proposed DS-GAN and compared approaches.
Values in the first column denote the target presentation of the corresponding approach, where $T$ means textual, and $V$ means visual.
It is observed that the proposed DS-GAN achieves the best performance in almost all graphic metrics as expected, for it benefits from CNN-LSTM, which helps recognize a pattern of sequences of geometric parameters, i.e., bounding boxes.
First of all, most elements in its generated layouts are valid, which is the basis.
Impressively, it does an excellent job of avoiding undesired overlap between elements, obtaining $Ove \downarrow$ only 37\% of CGL-GAN's.
It also has a good result on avoiding spatial non-alignment, reducing $Ali \downarrow$ by more than 25\%.
In terms of loose underlay effectiveness, DS-GAN is a little behind, but on the opposite, the leading in strict underlay effectiveness exactly proves it has not abused underlay elements.
As for content-aware metrics, both CGL-GAN and ours fall behind SmartText, showing there is indeed a long way for research to go toward content-aware visual-textual presentation layout.
Nevertheless, DS-GAN can achieve nearly comparable performance with CGL-GAN on avoiding unreadability.

\begin{figure*}[t]
  \vspace{0.4\baselineskip}  
  \centering
  {\renewcommand{\arraystretch}{0.2}
  \begin{tabular}{l@{\hspace{3.5pt}}c@{\hspace{1pt}}c@{\hspace{1pt}}c@{\hspace{1pt}}c@{\hspace{1pt}}c@{\hspace{1pt}}c@{\hspace{1pt}}c@{\hspace{1pt}}c}
  \raisebox{18pt}{\rotatebox{90}{Saliency map}} &
 \includegraphics[width=0.116\textwidth] {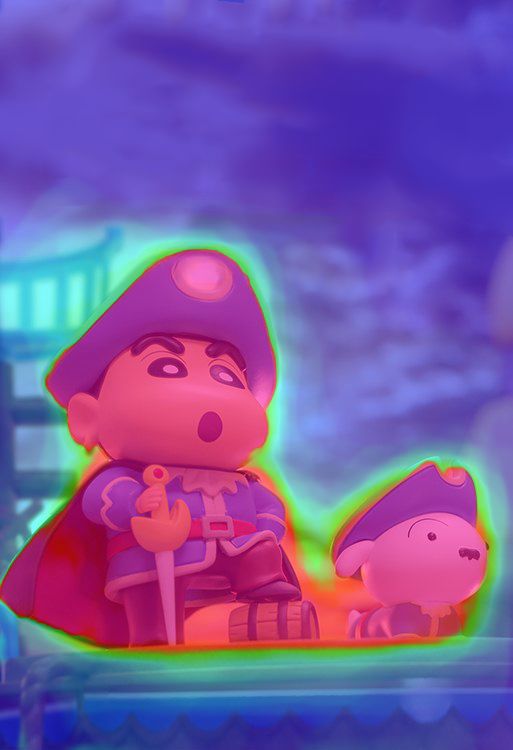} &
 \includegraphics[width=0.116\textwidth] {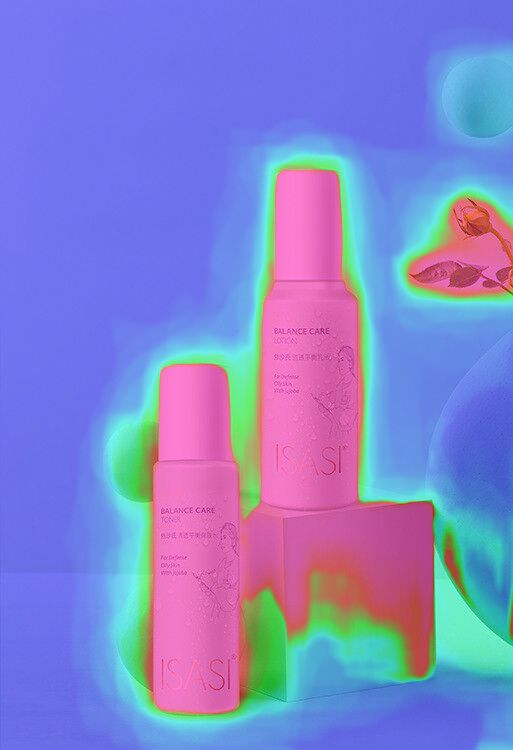} &
 \includegraphics[width=0.116\textwidth] {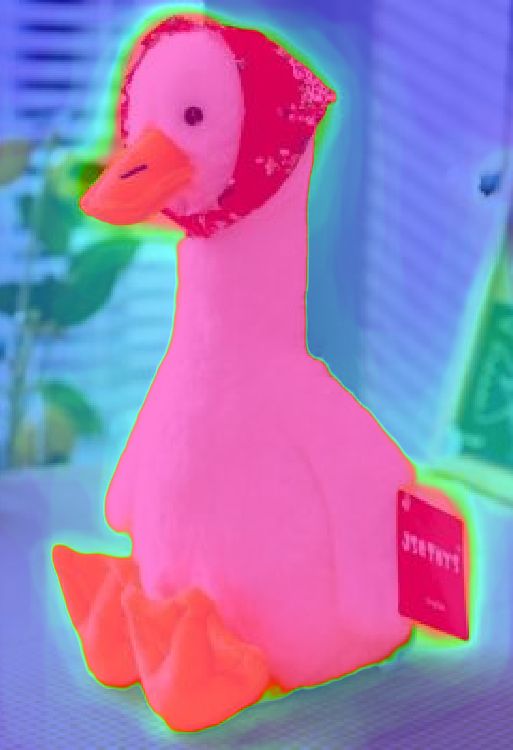} &
 \includegraphics[width=0.116\textwidth] {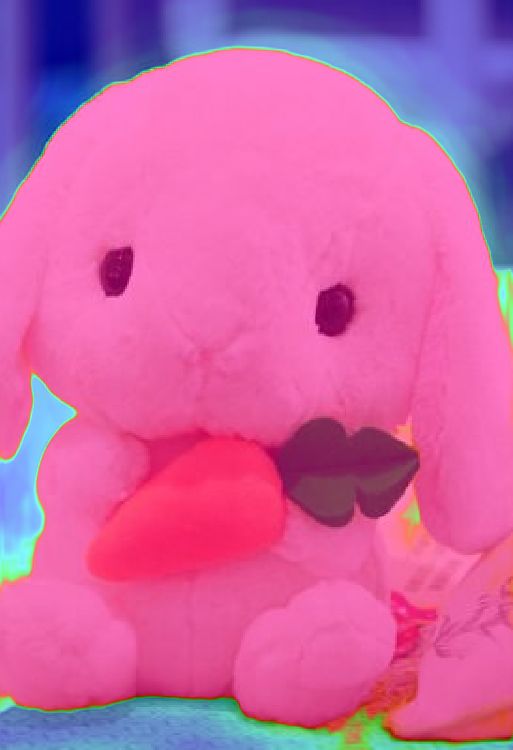} &
 \includegraphics[width=0.116\textwidth] {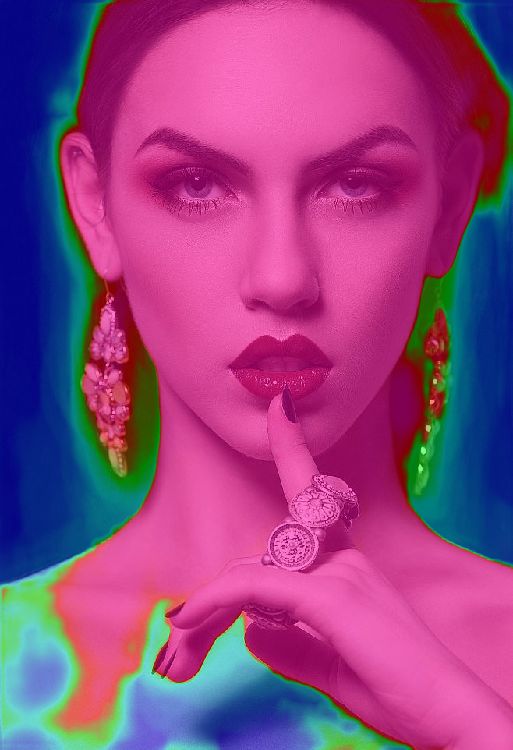} &
 \includegraphics[width=0.116\textwidth] {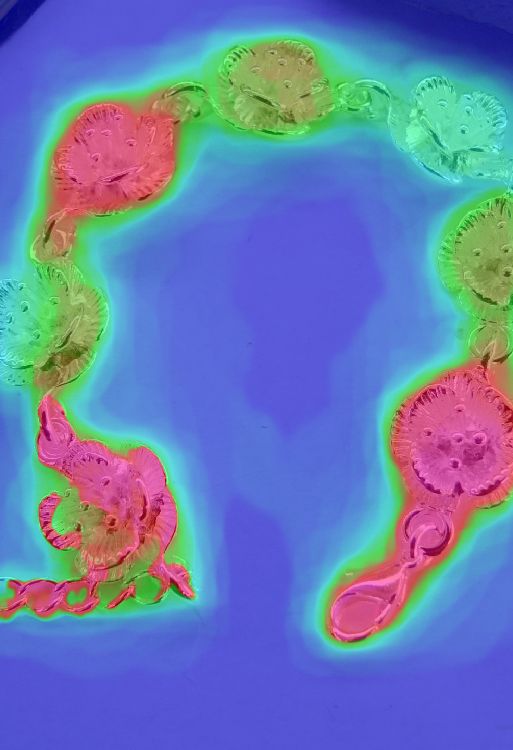} &
 \includegraphics[width=0.116\textwidth] {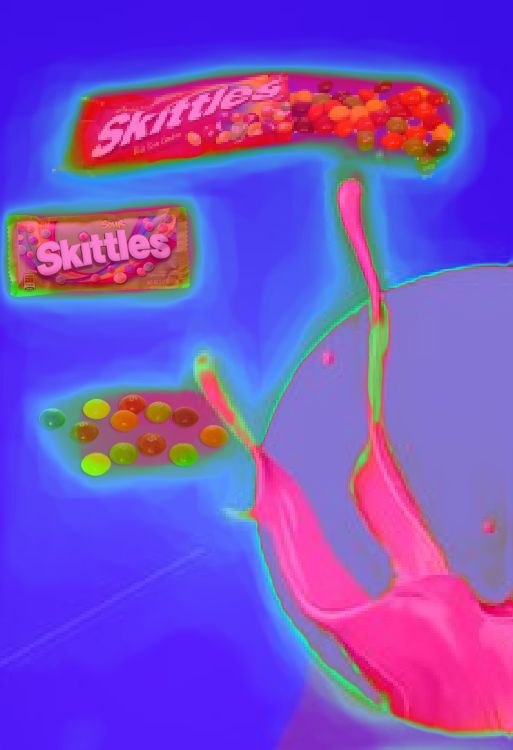} &
 \includegraphics[width=0.116\textwidth] {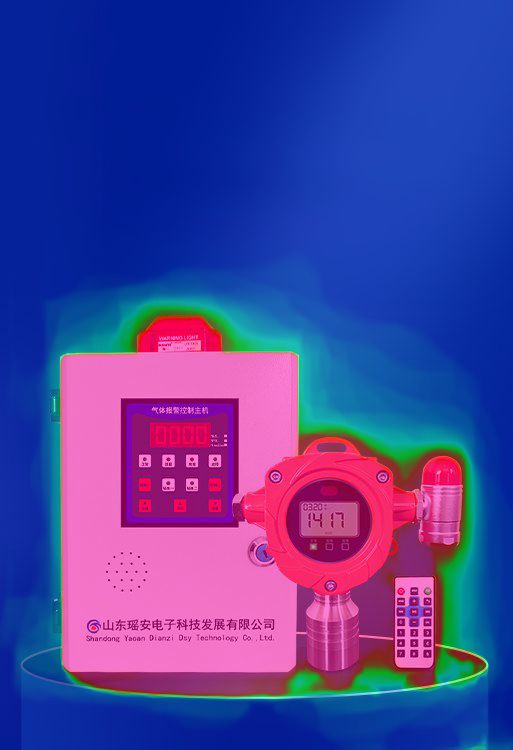}
 \\
\raisebox{22pt}{\rotatebox{90}{SmartText}} &
 \includegraphics[width=0.116\textwidth] {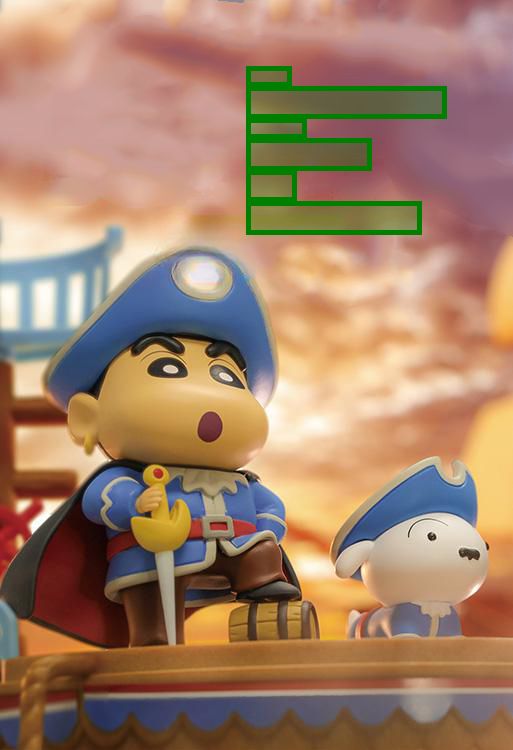} &
 \includegraphics[width=0.116\textwidth] {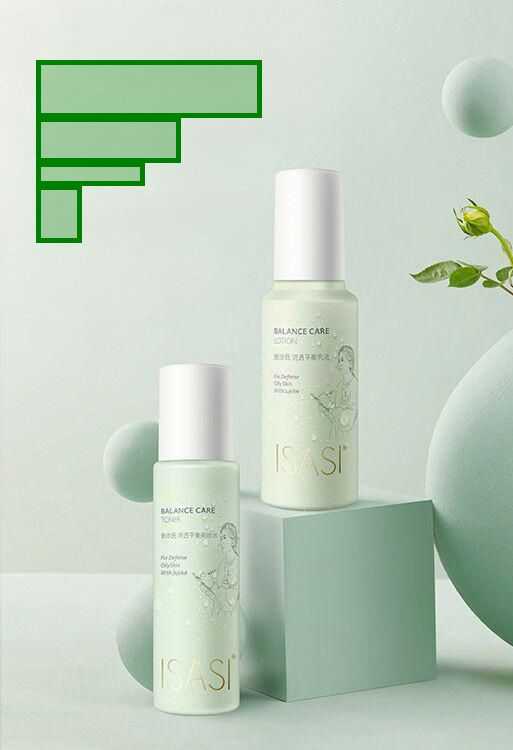} &
 \includegraphics[width=0.116\textwidth] {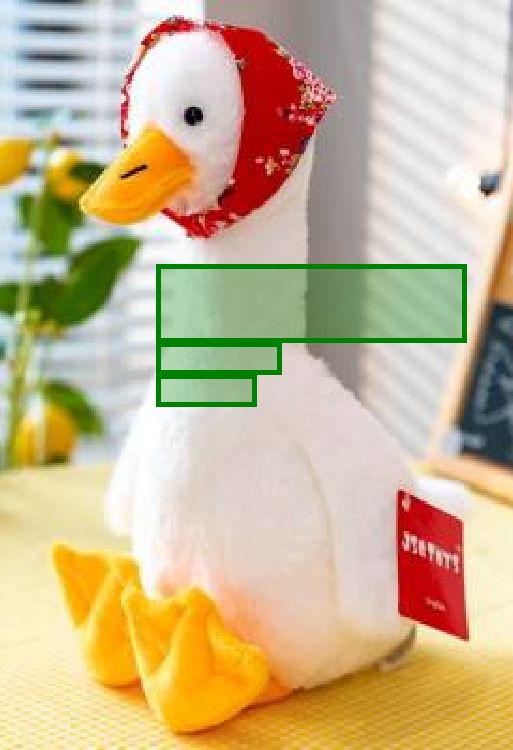} &
 \includegraphics[width=0.116\textwidth] {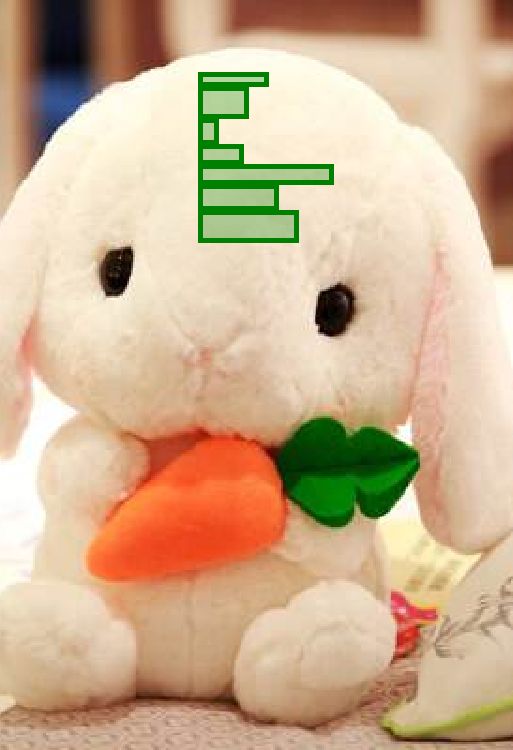} &
 \includegraphics[width=0.116\textwidth] {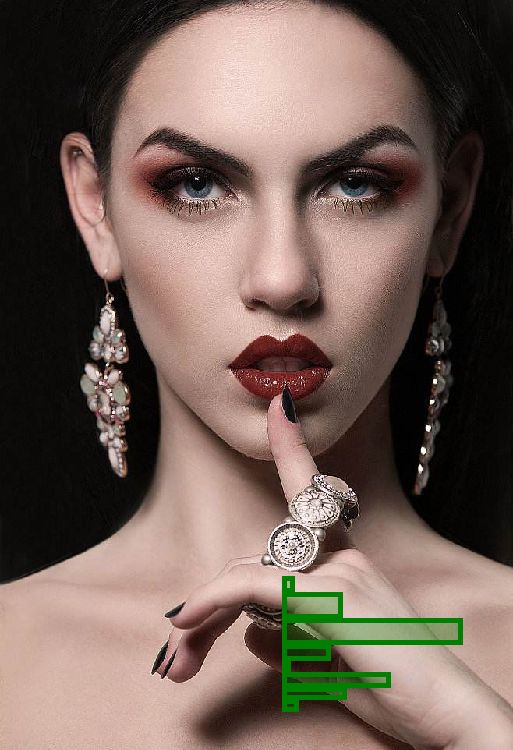} &
 \includegraphics[width=0.116\textwidth] {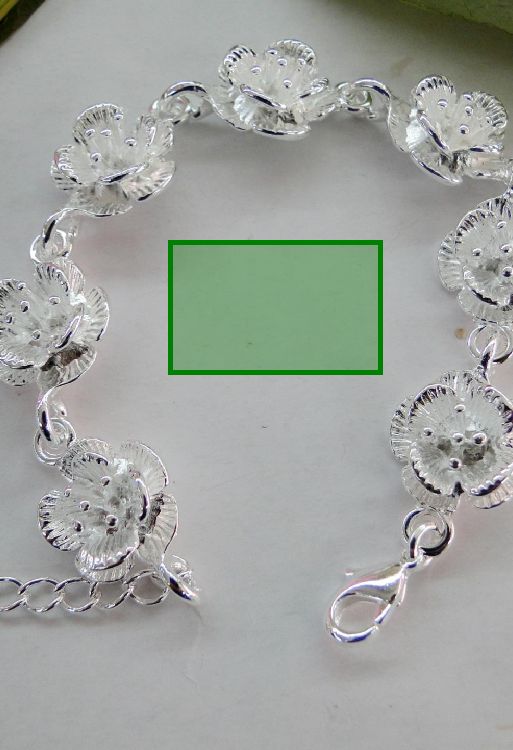} &
 \includegraphics[width=0.116\textwidth] {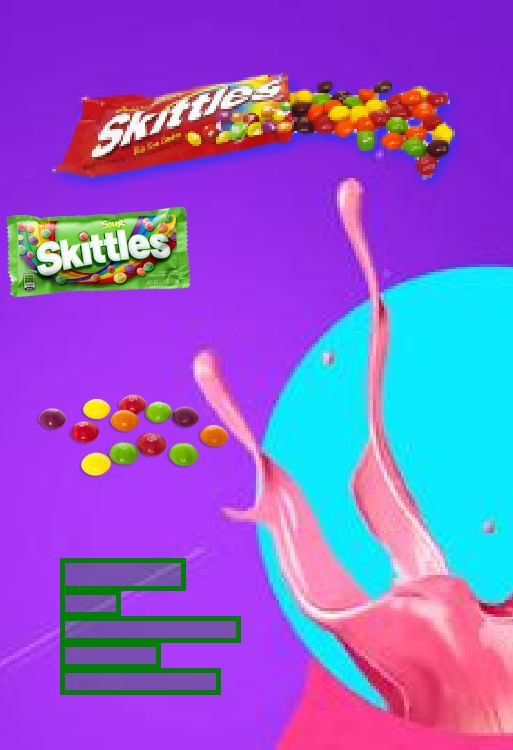} &
 \includegraphics[width=0.116\textwidth] {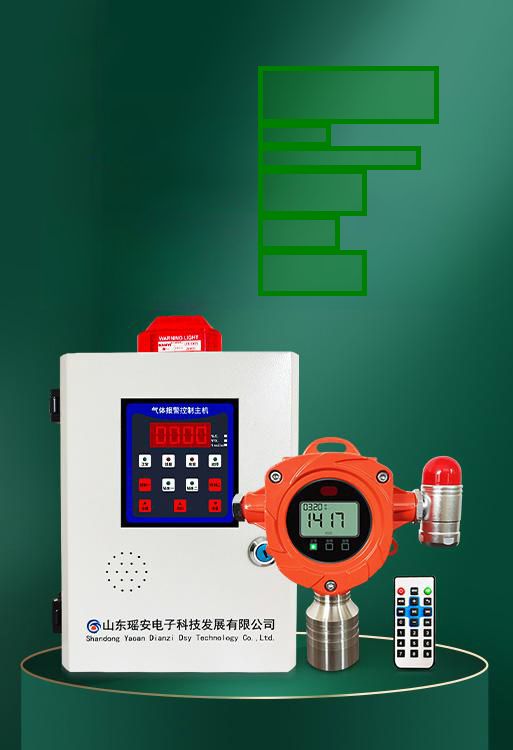}
 \\
 \raisebox{22pt}{\rotatebox{90}{CGL-GAN}} &
 \includegraphics[width=0.116\textwidth] {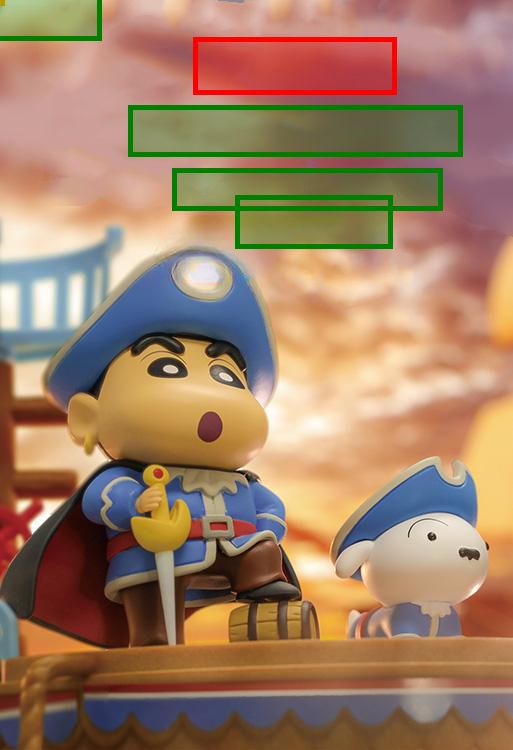} &
 \includegraphics[width=0.116\textwidth] {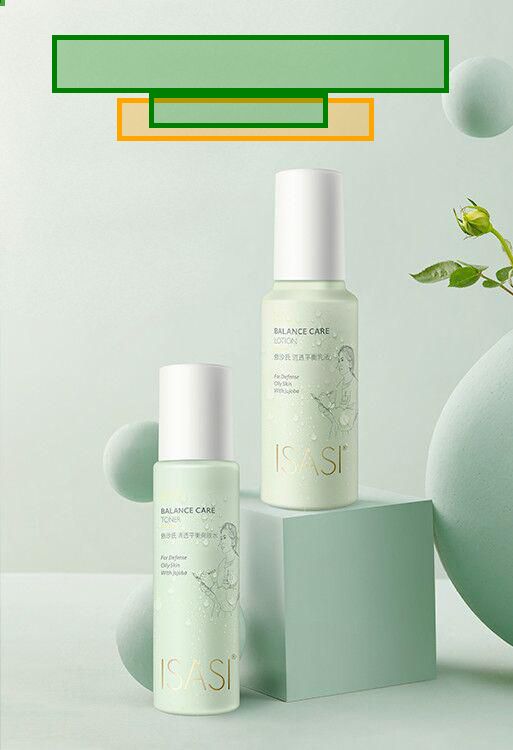} &
 \includegraphics[width=0.116\textwidth] {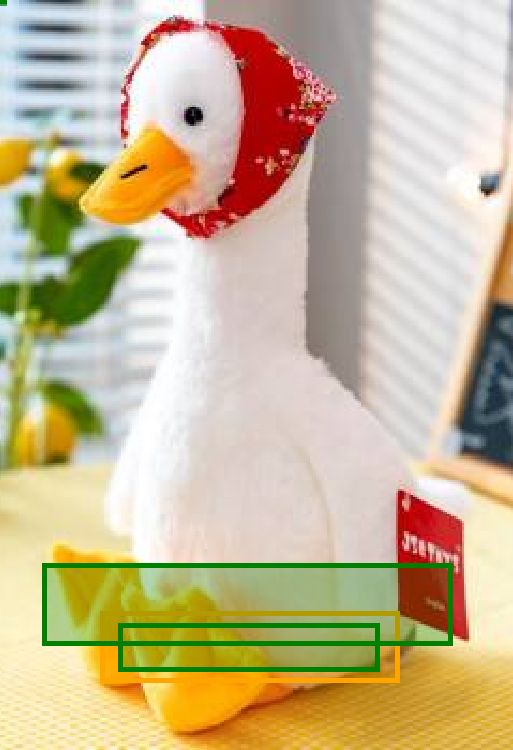} &
 \includegraphics[width=0.116\textwidth] {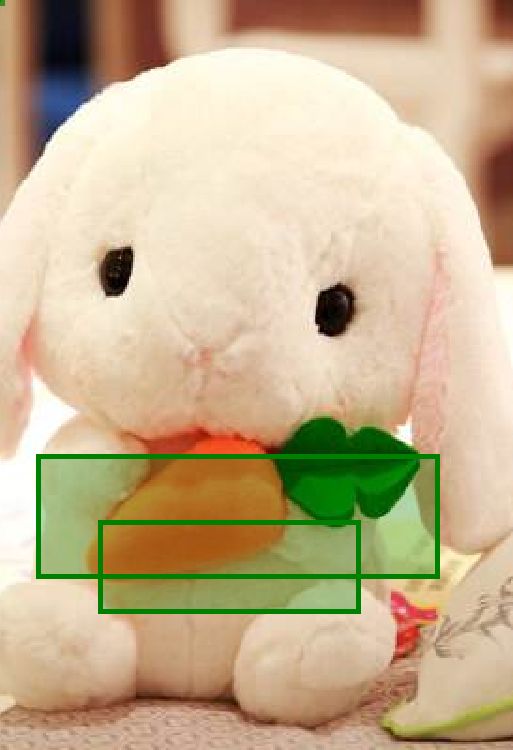} &
 \includegraphics[width=0.116\textwidth] {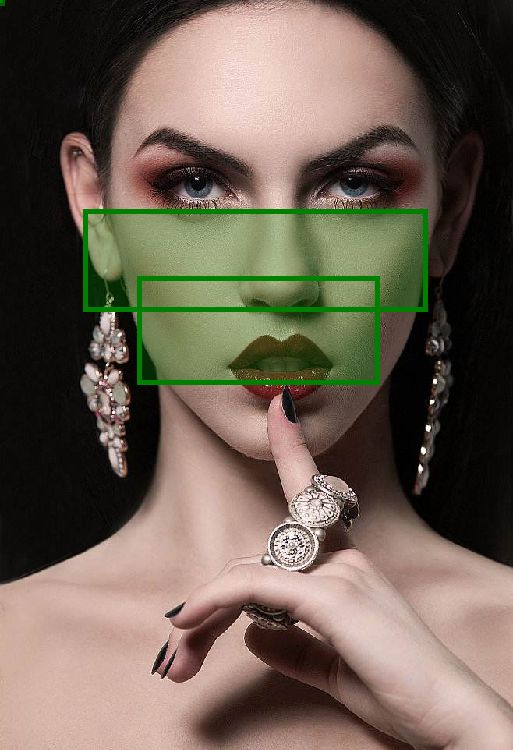} &
 \includegraphics[width=0.116\textwidth] {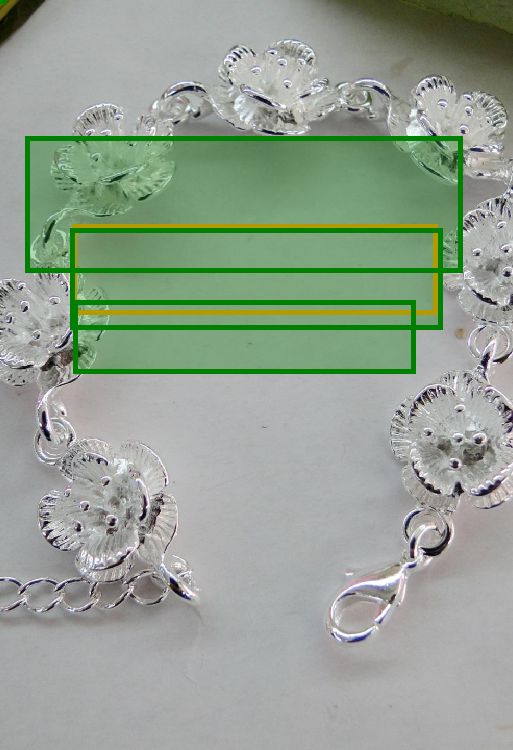} &
 \includegraphics[width=0.116\textwidth] {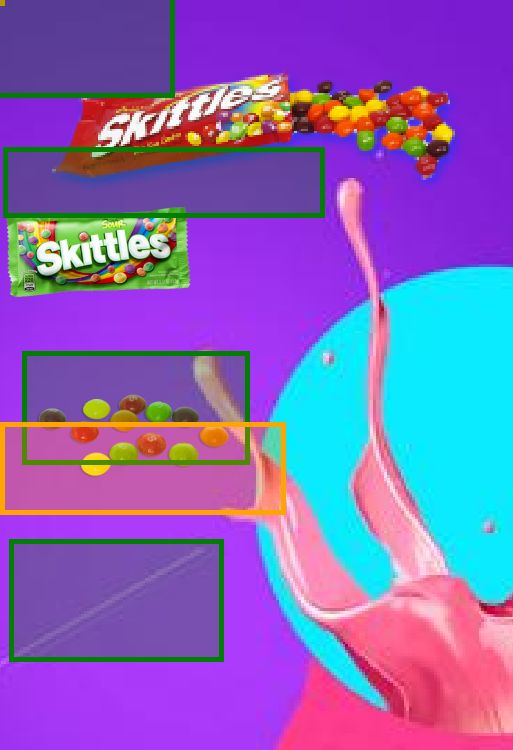} &
 \includegraphics[width=0.116\textwidth] {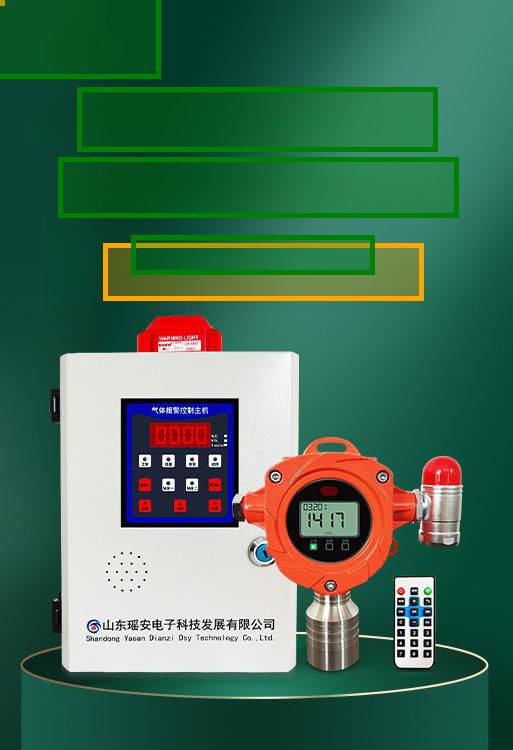}
 \\
 \raisebox{10pt}{\rotatebox{90}{DS-GAN (Ours)}} &
 \includegraphics[width=0.116\textwidth] {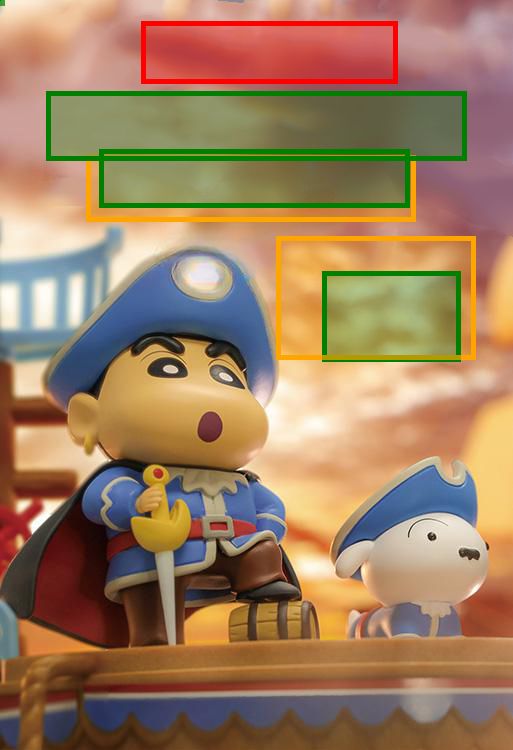} &
 \includegraphics[width=0.116\textwidth] {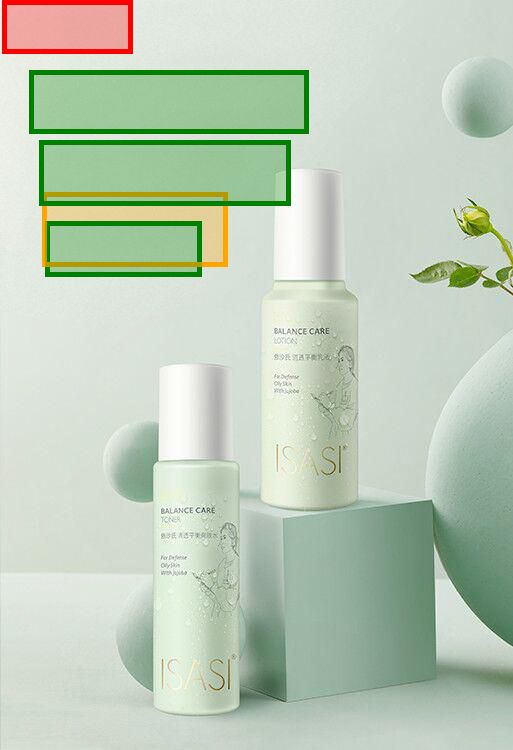} &
 \includegraphics[width=0.116\textwidth] {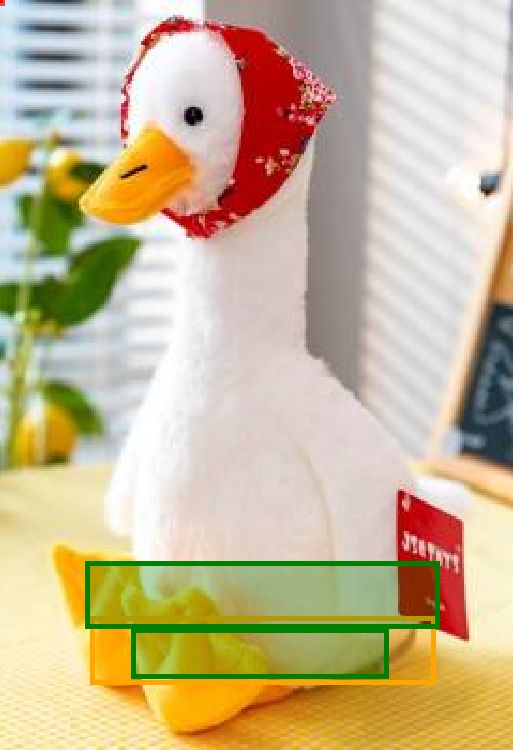} &
 \includegraphics[width=0.116\textwidth] {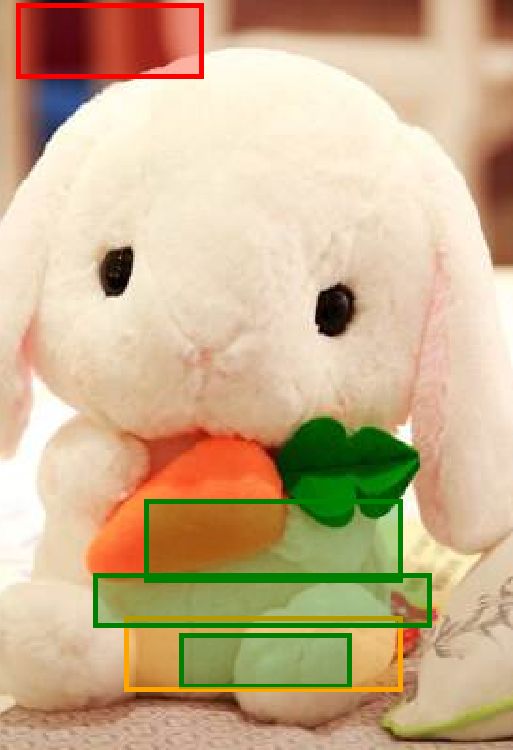} &
 \includegraphics[width=0.116\textwidth] {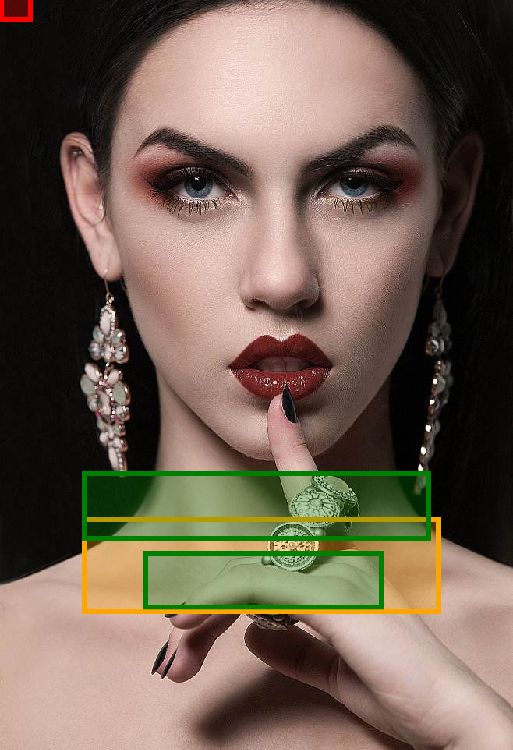} &
 \includegraphics[width=0.116\textwidth] {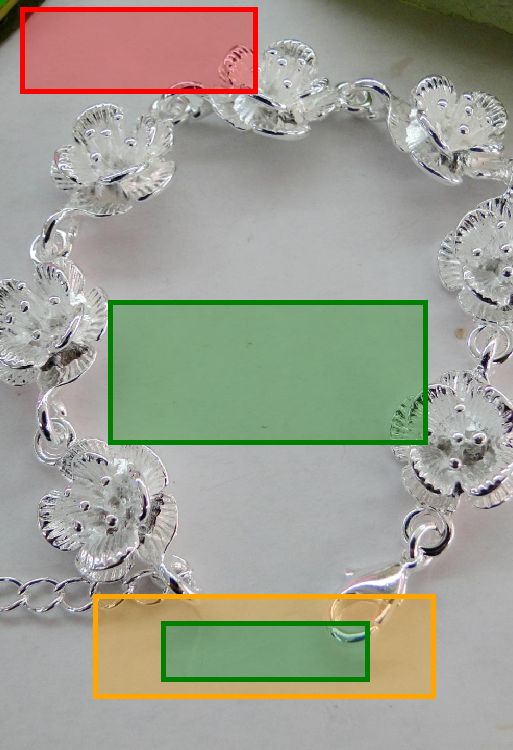} &
 \includegraphics[width=0.116\textwidth] {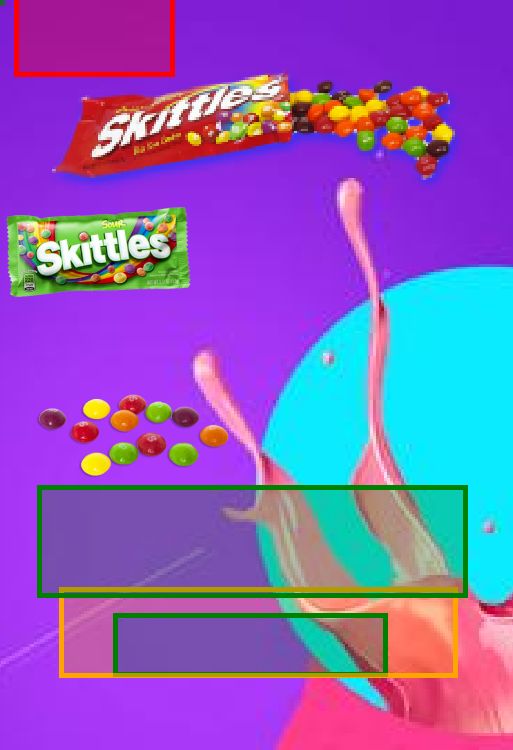} &
 \includegraphics[width=0.116\textwidth] {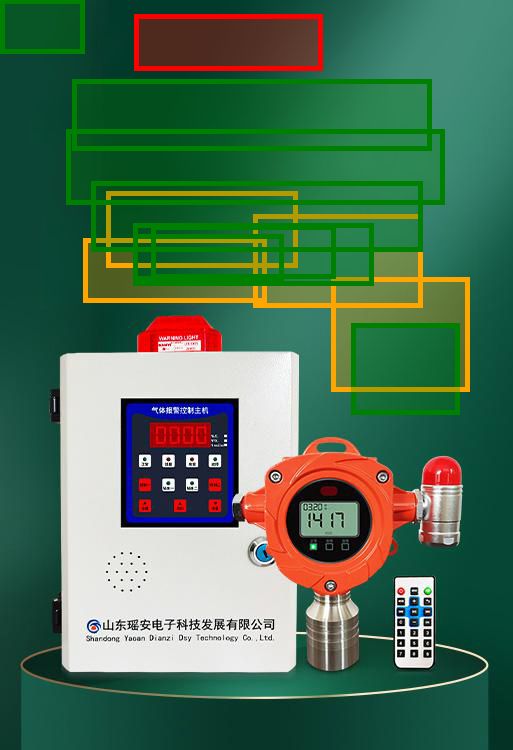} \\ \\
 & (a) & (b) & (c) & (d) & (e) & (f) & (g) & (h)

\end{tabular}
}
  \caption{Comparison of layouts generated by different approaches.}
  \label{fig:result}
    \vspace{-1\baselineskip}  
\end{figure*}

\cref{fig:result} shows several layouts generated from these approaches.
Visualized results verify the high $Uti \uparrow$ of DS-GAN, while cases like the one shown in column (b) fundamentally explain why it conversely gets behind in terms of $Occ \downarrow$ and $Rea \downarrow$.
That is, for its ability and inclination toward exploiting all available regions.
These cases tell that slight occlusion sometimes actually brings more appealing layouts. 
Moreover, DS-GAN's abilities to avoid overlay or non-alignment between elements are observed, and it can handle diverse canvases that annoy others.
The proposal-based method, i.e., SmartText, is problematic with cases shown in columns (c) and (d), where the only salient object almost fills the canvas.
CGL-GAN is prone to lose directions when confronting a canvas with a circle-outline object or numerous salient objects, as shown in columns (f) and (g).
Moreover, column (h) shows that DS-GAN actively generates more \emph{complex} layouts than others, even though some unpleasant overlay is witnessed. Since DS-GAN already does best on $Ove \downarrow$, further improvements can be challenging. We claim it as a promising direction, especially using our PKU PosterLayout dataset, which contains complex layouts.
Overall, the proposed approach generates more appealing layouts for diverse canvases by balancing its performance on graphics and content awareness.
Both quantitative and visualized results validate this conclusion.


\subsection{Ablation Study}
Since the CNN-LSTM model is the key to DS-GAN, an ablation experiment is conducted by remaining only the last fully connected layers to evaluate its effectiveness.
As shown in \cref{tab:ablation_body}, the monotonically decreasing performance strongly demonstrates the necessity of the model.
It is expected since CNN-LSTM helps behavior analysis, which is the main working logic of DS-GAN.

To gain insight into the effect of DSF, an ablation study is carried out.
Remember that an important capability of DSF is maintaining a descending order of element importance in the design sequence, which indicates that discarding the least important elements should be trivial to the final performance.
Therefore, the dependent variable is the length of the design sequences. We set it as (a) the maximum number of elements in all layout data or (b) 8, i.e., DS-GAN-8, and verify the effects on three different formation strategies, including (1) \emph{random} order, (2) ascending order of top-left coordinates, i.e., \emph{geometric}, and (3) the proposed \emph{DSF-based} order.
Experimental results are shown in \cref{tab:ablabtion}.
Except for the last column, the values represent the metrics under (b), and those in the parentheses are the difference between metrics under (a) and (b), calculated by (b) - (a) and thus show what happens if almost half of the elements are discarded.
Observing the results, we find the DSF-based gets the best performance as expected, for that random and geometric strategies may discard essential elements, especially the random one.
Moreover, by aggregating the total absolute difference, annotated as $AE \downarrow$, we verify that the perturbation brought by the number of elements is the most trivial when adopting the proposed DSF algorithm.
\section{Conclusion}
\label{sec:conclusion}
In this paper, we construct a new dataset and benchmark for content-aware visual-textual presentation layouts, named \textbf{PKU PosterLayout}.
With satisfactory layout variety, domain diversity, and content diversity, it is more challenging and expected to encourage further research.
We also propose a generative approach, \textbf{DS-GAN}, with the DSF algorithm to treat layout generation as a behavior modeling problem.
The DSF algorithm can form plain layout data into design sequences and help DS-GAN learn the pattern better.
Several experiments are conducted to verify
(1) the usefulness of the proposed benchmark and
(2) the effectiveness of the proposed approach that generates suitable layouts for diverse canvases.

The future works mainly lie in two aspects:
(1) Further improving content-aware performance without violating graphic performance, which may be done by replacing the off-the-shelf saliency detection method with a dedicated one or involving it in the end-to-end training process.
(2) Devoted to high-quality complex layout generation, which is promising and achievable utilizing the first public dataset containing complex layouts-- PKU PosterLayout.

\section*{Acknowledgements}
This work was supported by grants from the National Natural Science Foundation of China (61925201, 62132001, 62272013, U22B2048) and Meituan.

{\small
\bibliographystyle{ieee_fullname}
\bibliography{PaperForReview}
}

\end{document}